\documentclass[final,3p,times,authoryear]{elsarticle}

\usepackage{amssymb}

\usepackage{hyperref}
\usepackage{amsmath}
\usepackage{makecell}
\usepackage{multirow}
\usepackage{longtable}
\usepackage{boldline}
\usepackage{subfig}
\usepackage{eucal}
\usepackage[ruled]{algorithm2e}

\journal{Computer Speech and Language}

\begin{document}

\begin{frontmatter}



\title{DORA: Towards Policy Optimization for Task-oriented Dialogue System with Efficient Context}


\author[inst1]{Hyunmin Jeon}

\affiliation[inst1]{organization={Computer Science and Engineering},
            addressline={Pohang University of Science and Technology}, 
            country={South Korea}}

\author[inst1,inst2]{Gary Geunbae Lee \corref{cor1}}

\affiliation[inst2]{organization={Graduate School of Artificial Intelligence},
            addressline={Pohang University of Science and Technology}, 
            country={South Korea}}

\begin{abstract}
\let\thefootnote\relax\footnotetext{E-mail: jhm9507@postech.ac.kr (Jeon), gblee@postech.ac.kr (Lee).}
Recently, reinforcement learning (RL) has been applied to task-oriented dialogue systems by using latent actions to solve shortcomings of supervised learning (SL).
In this paper, we propose a multi-domain task-oriented dialogue system, called \textbf{D}ialogue System with \textbf{O}ptimizing a \textbf{R}ecurrent \textbf{A}ction Policy using Efficient Context (DORA), that uses SL, with subsequently applied RL to optimize dialogue systems using a recurrent dialogue policy.
This dialogue policy recurrently generates explicit system actions as a both word-level and high-level policy.
As a result, DORA is clearly optimized during both SL and RL steps by using an explicit system action policy that considers an efficient context instead of the entire dialogue history.
The system actions are both interpretable and controllable, whereas the latent actions are not.
DORA improved the success rate by 6.6 points on MultiWOZ 2.0 and by 10.9 points on MultiWOZ 2.1.
\end{abstract}



\begin{keyword}
Task-oriented dialogue system \sep Multi-domain dialogue \sep Policy optimization \sep Recurrent Action Policy \sep Efficient context
\end{keyword}

\end{frontmatter}


\section{Introduction}
\label{sec:introduction}
Task-oriented dialogue systems are designed to use conversations to help users achieve goals in specific domains.
In multi-domain dialogues, users have goals across multiple domains, so to respond adequately, the systems should track the flow of conversation among domains.
This necessity complicates development of task-oriented dialogue systems that can process multi-domain dialogues.

Typical dialogue systems have a pipeline architecture that consists of a natural language understanding (NLU) module, belief tracker, dialogue policy, and natural language generation (NLG) module.
Recent work has focused on neural dialogue systems that are end-to-end trainable by using supervised learning (SL) \citep{zhang2020task, hosseini2020simple, peng2020soloist, yang2020ubar}.
SL is an efficient method to train neural networks, but use of SL to train task-oriented dialogue systems has some limitations.
First, training using SL requires annotations, which can be erroneous.
Furthermore, task-oriented dialogues do not have a definite optimal answer; many options must be considered when formulating an answer to a user utterance.
Thus, the systems can become biased by the annotations.
To address these problems, several approaches have attempted to train dialogue systems by using reinforcement learning (RL) to optimize dialogue policy that samples latent variables as actions \citep{zhao2019rethinking, wang2020modelling, lubis2020lava, lee2020sumbt+}.

Even though use of latent actions has improved the success rate of dialogue systems, the strategy has several limitations in task-oriented dialogue systems.
Training the systems from scratch using RL is almost impossible, so they are generally pre-trained using SL.
However, latent actions have no gold-standard outputs; therefore, the NLG module, which follows the dialogue policy, cannot be given clear prior information for response generation during the SL step.
Furthermore, humans cannot interpret latent actions; this weakness complicates the task of determining the appropriate action space of latent variables, and of judging whether the latent variables represent this space well.

Another drawback of previous methods is that they use the entire dialogue history as the system input.
Use of large-scale pre-trained language models, such as BERT \citep{devlin2019bert} and GPT-2 \citep{radfordlanguage}, has been a trend in NLP field and has improved the success rate of dialogue systems, but it significantly increases the model size.
Even with the same increase in input length, the memory usage increases faster on larger model, and the dialogue history lengthens as the conversation progresses, so use of the dialogue history increases training cost.

\begin{figure}
    \centering
	\includegraphics[width=0.9\textwidth]{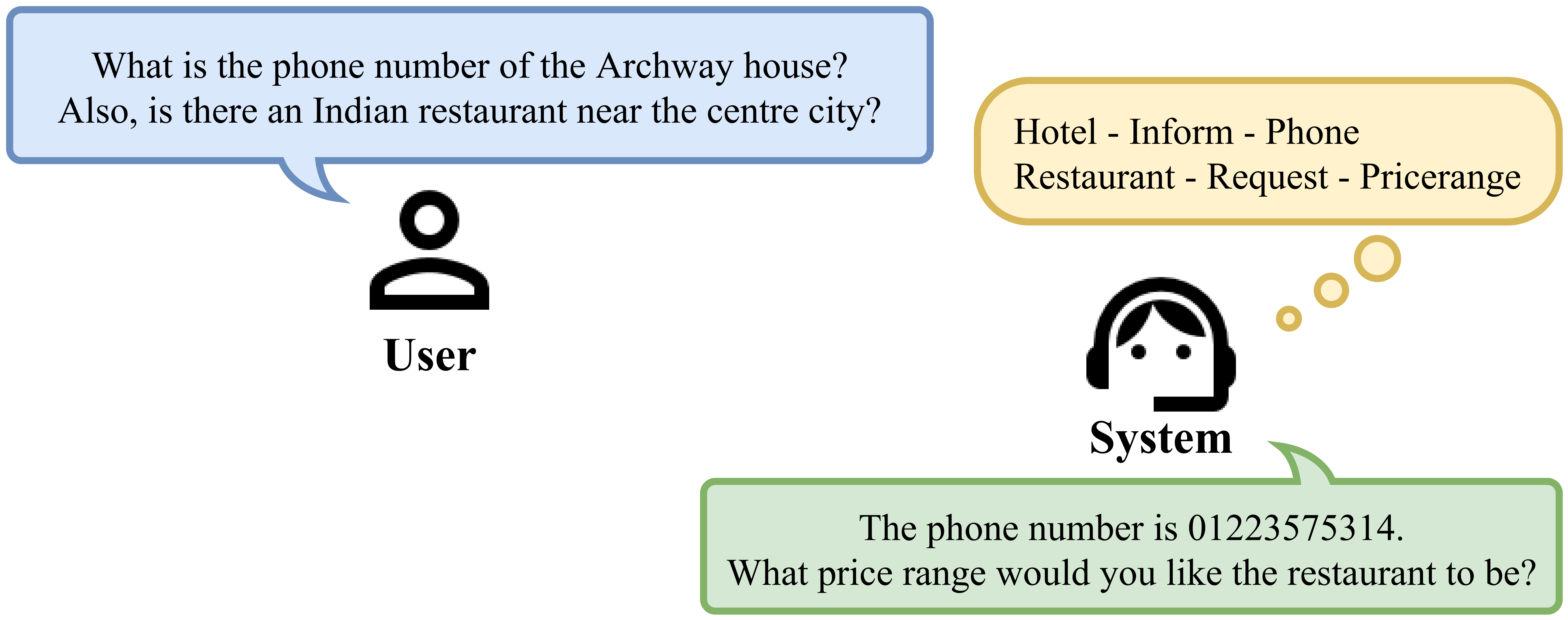}
    \caption{Two system actions across two domains and a corresponding system response to respond to a complex utterance in a multi-domain dialogue.}
    \label{figure1}
\end{figure}

In this study, we propose \textbf{D}ialogue System with \textbf{O}ptimizing a \textbf{R}ecurrent \textbf{A}ction Policy using Efficient Context (DORA), a task-oriented dialogue system that uses explicit system actions instead of latent actions, and summarized input instead of dialogue history, to address the above limitations.
In multi-domain dialogues, systems should generate multiple system actions at once to respond to complex user utterances.
DORA processes this by recurrently generating system actions, not just choosing one action.
Figure \ref{figure1} shows an example of this process.

Use of explicit system actions instead of latent actions enables clear optimization of an NLG module given gold-standard system actions even when the dialogue policy is not optimized, during the SL step.
Furthermore, the system actions are obviously interpretable.
Thus, the generated system actions can be used for reward shaping during the RL step, and the optimized dialogue policy can be controlled by post-processing for various purposes.

DORA only uses a current user utterance instead of the entire dialogue history.
However, abandoning dialogue history causes losses of contextual information from previous turns.
To prevent the losses, DORA also uses the domain state and belief state as input.
The domain state and belief state are updated as the conversation progresses, so they can efficiently represent contextual information that is accumulated during multiple turns.
Removing dialogue history makes the memory usage almost constant regardless of the length of conversations; therefore, DORA can be stably and efficiently trained even with long conversations.

We evaluated DORA on MultiWOZ 2.0 \citep{budzianowski2018multiwoz} and MultiWOZ 2.1 \citep{eric2020multiwoz}, which are standard benchmark datasets for multi-domain task-oriented dialogue systems.
DORA achieved higher success rate than previous methods.
The results demonstrate that use of explicit system actions enables clear optimization of the system during both the SL and RL steps, and that the input context using the domain state and belief state can efficiently perform the role of dialogue history.

\begin{figure}[ht]
    \centering
	\includegraphics[width=0.9\textwidth]{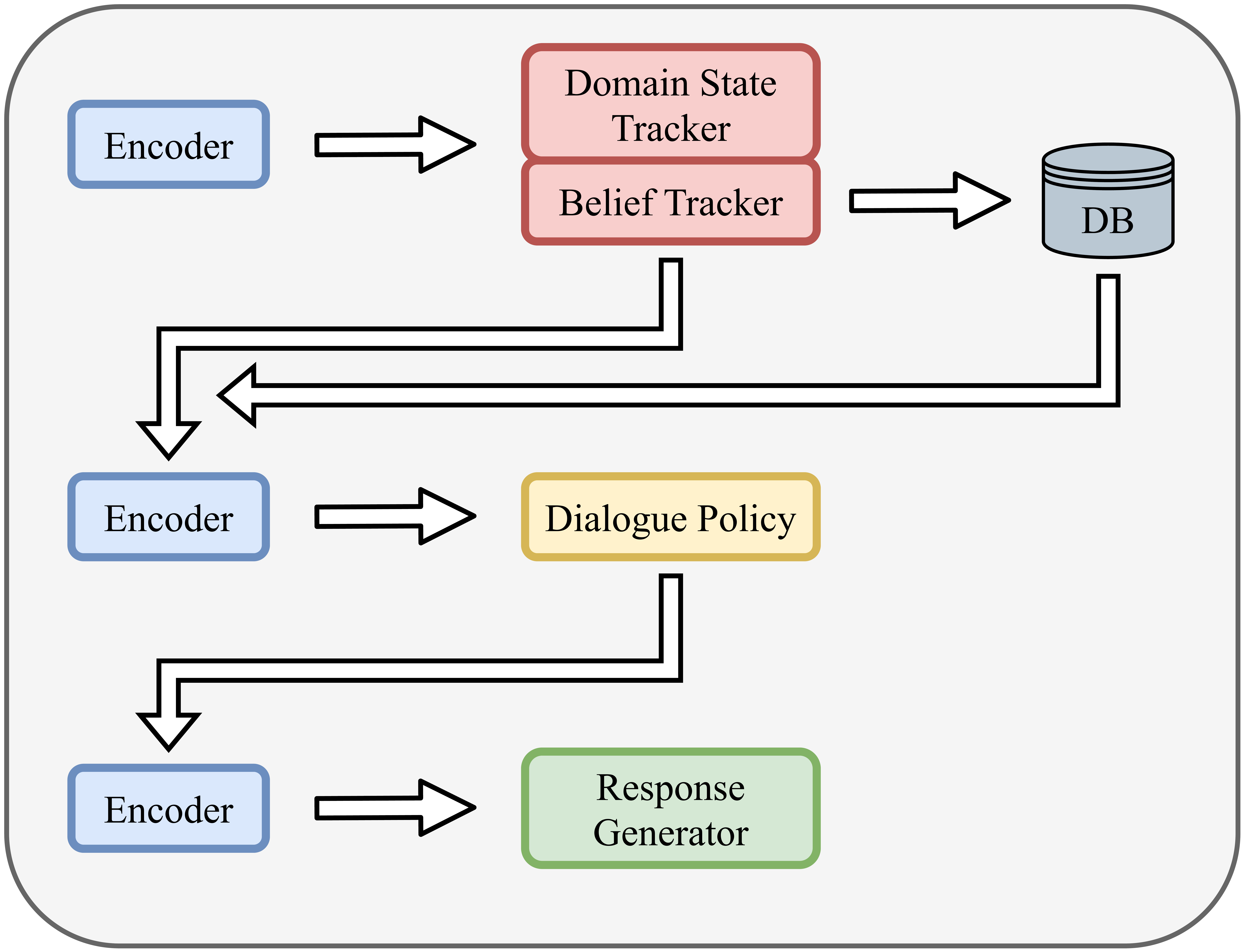}
    \caption{Architecture overview of DORA: a neural network consisting of three sequence-to-sequence sub networks with a shared encoder and a task-specific DB.}
    \label{figure2}
\end{figure}

\section{Related Work}
\label{sec:related:works}
Recent work on task-oriented dialogue systems has focused on building end-to-end trainable dialogue systems in multi-domain dialogues \citep{zhang2020task, hosseini2020simple, peng2020soloist, lin2020mintl, yang2020ubar, zhang2020probabilistic}.
Some researchers have attempted to use large-scale pre-trained language models, such as BERT and GPT-2, to transfer abundant contextual representation trained on vast corpus.
BERT performs well as an NLU module by encoding the dialogue history \citep{lee2020sumbt+}.
Some methods consider task-oriented dialogues as language modeling tasks using GPT-2 \citep{hosseini2020simple, peng2020soloist, yang2020ubar}.
Pre-trained sequence-to-sequence models, such as BART \citep{lewis2020bart} and T5 \citep{raffel2019exploring}, also can be used for task-oriented dialogue systems.
BART and T5 have been evaluated as backbone models for multi-domain task-oriented dialogues \citep{lin2020mintl}.

To address the limitations of SL for task-oriented dialogue systems, various approaches have used RL.
To apply RL in dialogue systems, a policy sampling actions given a state should be included in the systems.
The na\"ive approach is to treat an NLG module as a low-level policy and to use words as actions \citep{lewis2017deal, das2017learning, kottur2017natural}.
However, the low-level and word-level policies have large action space, so optimization during the RL step is difficult and inefficient.
Therefore, recent studies have considered latent-level policies that sample latent variables, instead of words, as actions \citep{zhao2019rethinking, wang2020modelling, lubis2020lava, lee2020sumbt+}.
Optimization of latent action policies as high-level policies have improved the success rate of task-oriented dialogue systems.

To reduce the burdens of using the entire dialogue history, several methods to efficiently represent the history have been proposed.
One encodes utterances and efficiently accumulates latent representations of the history as the conversation progresses \citep{gupta2018efficient}.
Another presents an efficient belief tracking system by using the previous belief state as an additional input instead of the entire history \citep{kim2020efficient}.

In our work, we demonstrate a method that solves both the shortcomings of latent action policies and the inefficiency of dialogue history.

\section{Dialogue System with Optimizing a Recurrent Action Policy using Efficient Context}
\label{sec:method}

In this section, we describe the architecture (Figure \ref{figure2}) of DORA, and then we explain how to optimize the system and to construct the efficient context for system input.
In every turn, the neural network sequentially predicts the domain state and belief state, makes system actions, and generates a system response.

\begin{figure}
    \centering
	\includegraphics[width=0.9\textwidth]{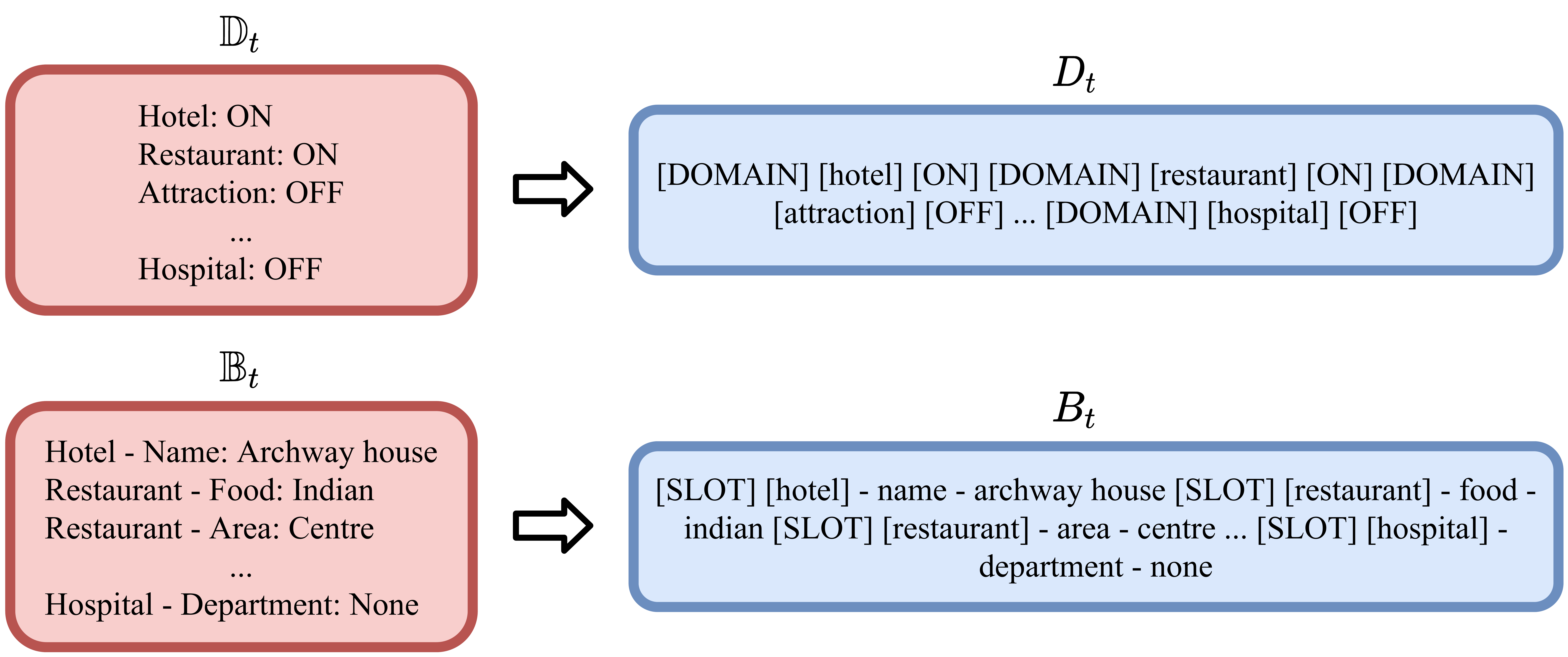}
    \caption{Conversion processes of the domain state and belief state from sets to sentences. Words enclosed in brackets are special words for distinguishing from general words.}
    \label{figure3}
\end{figure}

\subsection{Context Construction}
\label{subsec:context}
The belief state indicates the constraints of user goals; it consists of various predefined slots and their corresponding values.
Tracking the belief state is an essential task for task-oriented dialogue systems that should catch user goals during long conversations.
In addition to the belief state, DORA tracks the domain state, which indicates whether each domain is activated in the current conversation.
By tracking the domain state, the system can trace the flow of conversation from a domain perspective, thereby simplifying the task of understanding the purpose of a user in multi-domain dialogues.

\begin{figure}
    \centering
	\includegraphics[width=0.9\textwidth]{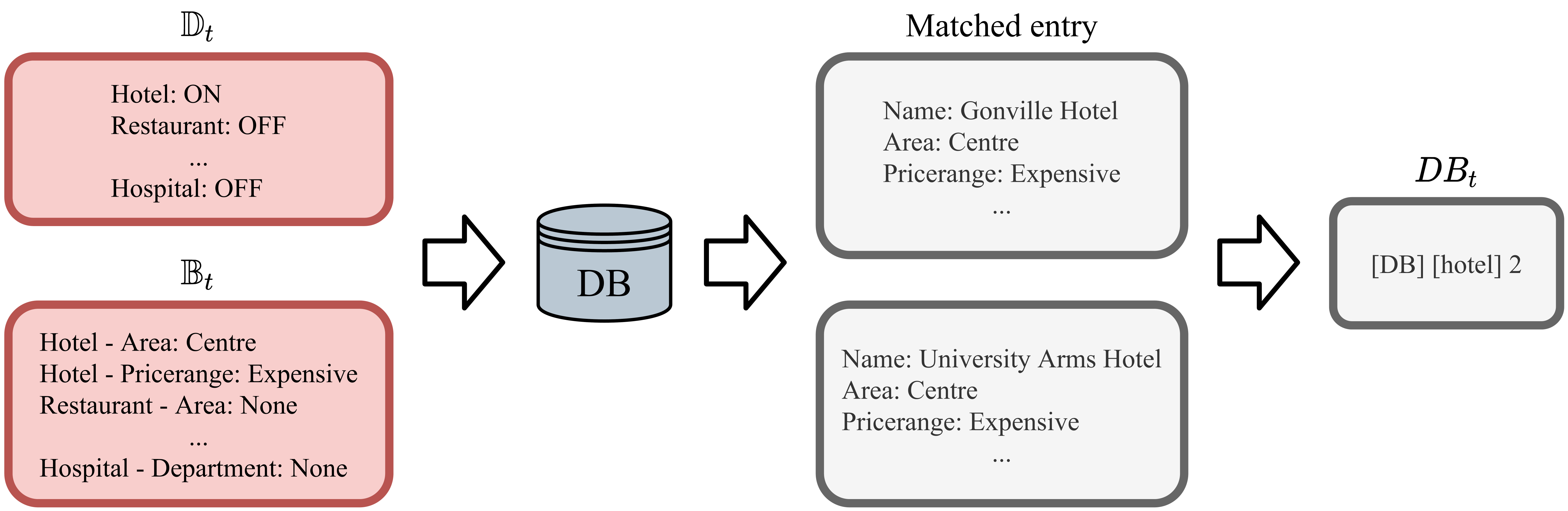}
    \caption{Process of DB search to obtain $DB_t$.}
    \label{figure4}
\end{figure}

Instead of using the entire dialogue history, we efficiently construct an input context using the domain state and belief state in addition to user utterance.
The context changes over three steps of domain state and belief tracking, system action generation, and system response generation.
The initial context $C_t^{Init}$ for domain state and belief tracking on turn $t$ consists of the current user utterance $U_t$, the previous domain state $\mathbb{D}_{t-1}$, and the previous belief state $\mathbb{B}_{t-1}$.
On current turn $t$, $\mathbb{D}_{t-1}$ and $\mathbb{B}_{t-1}$ have contextual information that had accumulated until the previous turn, and they can perform as system inputs instead of the dialogue history.
$\mathbb{D}_{t-1}$ is a set of binary values that indicates whether each domain is activated.
$\mathbb{B}_{t-1}$ is a set of values for each slot.
To feed the domain state and belief state into the encoder, we convert $\mathbb{D}_{t-1}$ to $D_{t-1}$ and $\mathbb{B}_{t-1}$ to $B_{t-1}$, in sentence forms by using some special words (Figure \ref{figure3}).
$C_t^{Init}$ is represented as
\begin{align}
    \label{equation1}
    \begin{split}
        C_t^{Init} &= \texttt{[CLS]} \oplus U_t \oplus \texttt{[SEP]} \oplus D_{t-1} \oplus B_{t-1} \oplus \texttt{[SEP]}
        = \left\{ c_{t, 1}^{Init}, \cdots, c_{t, \left| C_t^{Init} \right|}^{Init} \right\}
    \end{split},
\end{align}
where special word \texttt{[CLS]} represents the entire sentence and \texttt{[SEP]} distinguishes between the utterance and states; $\oplus$ indicates concatenation.
By considering the current domain state $\mathbb{D}_t$ and belief state $\mathbb{B}_t$, the DB operator queries a task-specific DB to obtain a list of matched entries, and converts the list to sentence form $DB_t$ (Figure \ref{figure4}).
The belief context $C_t^{Belief}$ for system action generation includes $D_t$ and $B_t$ instead of $D_{t-1}$ and $B_{t-1}$, respectively, and additionally includes $DB_t$.
$C_t^{Belief}$ is represented as follows:
\begin{align}
    \label{equation2}
    \begin{split}
        C_t^{Belief} &= \texttt{[CLS]} \oplus U_t \oplus \texttt{[SEP]} \oplus D_t \oplus B_t \oplus DB_t \oplus \texttt{[SEP]} = \left\{ c_{t, 1}^{Belief}, \cdots, c_{t, \left| C_t^{Belief} \right|}^{Belief} \right\}
    \end{split}.
\end{align}
The action context $C_t^{Act}$ for system response generation additionally contains the generated system actions $A_t$.
$C_t^{Act}$ is represented as follows:
\begin{align}
    \label{equation3}
    \begin{split}
        C_t^{Act} &= \texttt{[CLS]} \oplus U_t \oplus \texttt{[SEP]} \oplus D_t \oplus B_t \oplus DB_t \oplus A_t \oplus \texttt{[SEP]} = \left\{ c_{t, 1}^{Act}, \cdots, c_{t, \left| C_t^{Act} \right|}^{Act} \right\}
    \end{split}.
\end{align}

\subsection{Model Formulation}
\label{subsec:formulation}
We use pre-trained BERT for the shared encoder.
$C_t^{Init}$, $C_t^{Belief}$, and $C_t^{Act}$ are fed into the shared encoder to perform sequential tasks of domain state and belief tracking, system action generation, and system response generation, respectively.

\subsubsection{Domain State Tracker}
\label{subsubsec:domain:state:tracker}
BERT encodes $C_t^{Init}$ to vector representations $H_t^{Init}$, and then $h_{t, 1}^{Init}$ that represents the \texttt{[CLS]} word is fed into a fully connected (FC) layer $W_{pool} \in \mathbb{R}^{dim_H \times dim_H}$.
Then, a \texttt{tanh} activation function is applied to obtain a pooled output $o_t^{Init}$ as
\begin{align}
    \label{equation4} H_t^{Init} &= \texttt{BERT}\left( C_t^{Init} \right) = \left\{ h_{t, 1}^{Init}, \cdots, h_{t, \left| C_t^{Init} \right|}^{Init} \right\} \in \mathbb{R}^{\left| C_t^{Init} \right| \times dim_H}, \\
    \label{equation5} o_t^{Init} &= \texttt{tanh}\left( W_{pool}h_{t, 1}^{Init} \right) \in \mathbb{R}^{dim_H},
\end{align}
where $dim_H$ is the hidden size.
We use an FC layer $W_{domain} \in \mathbb{R}^{1 \times dim_H}$ for the domain state tracker, followed by a sigmoid activation function $\sigma$.
$D_{t-1}$ contains \texttt{[DOMAIN]}, a special word for representing each domain, and the vector representations of \texttt{[DOMAIN]} are used to predict the current domain state $D_t$ as
\begin{align}
    \label{equation6} H_t^{Domain} &= \left\{ \sigma\left( W_{domain} h_{t, i}^{Init} \right) \Big\vert c_{t, i}^{Init} = \texttt{[DOMAIN]}\right\} \in \mathbb{R}^{N_D}, \\
    \label{equation7} \mathbb{D}_t &= \left\{ \!\!\! \begin{array}{l}
        \mathbb{D}_t^d \Big\vert h_{t, d}^{Domain} \in H_t^{Domain}, \quad \mathbb{D}_t^d = \left\langle \begin{array}{ll}
            ON & \text{if } h_{t, d}^{Domain} \geq 0.5 \\
            OFF & \text{else}
        \end{array} \right.
    \end{array} \!\!\!\!\!\! \right\},
\end{align}
where $d \in [1, N_D]$ is domain index, and $N_D$ is the number of domains; i.e., a constant that indicates the number of \texttt{[DOMAIN]} in $D_{t-1}$.

\subsubsection{Belief Tracker}
\label{subsubsec:belief:tracekr}
Belief tracking is divided into two steps: slot-gate prediction and slot-value generation.
The number of slots varies according to the slot design.
However, predicting all slot-values every turn is inefficient because only some slots are activated at a time, generally.
Therefore, we predict slot-gates that indicates whether each slot is activated in current conversation, and then we generate slot-values for activated slots.
We use an FC layer $W_{gate} \in \mathbb{R}^{4 \times dim_H}$ for slot-gate prediction and a GRU \citep{cho2014properties} for slot-value generation.
$B_{t-1}$ contains \texttt{[SLOT]}, a special word to represent each slot, and the vector representations of \texttt{[SLOT]} are used for slot-gate prediction and slot-value generation.
The current slot-gate $G_t$ is classified as
\begin{align}
    \label{equation8} H_t^{Slot} &= \left\{ h_{t, i}^{Init} \Big\vert c_{t, i}^{Init} = \texttt{[SLOT]} \right\} \in \mathbb{R}^{N_S \times dim_H}, \\
    \label{equation9} G_t &= \left\{ g_t^s \Big\vert h_{t, s}^{Slot} \in H_t^{Slot}, \quad g_t^s = \texttt{argmax}\left( \texttt{softmax}\left( W_{gate} h_{t, s}^{Slot} \right) \right) \right\},
\end{align}
where $s \in [1, N_S]$ is slot index, and $N_S$ is the number of slots; i.e., a constant that indicates the number of \texttt{[SLOT]} in $B_{t-1}$.
The slot-gates have four classes: $g_t^s \in \left\{ \texttt{Update}, \texttt{Copy}, \texttt{Dontcare}, \texttt{Delete} \right\}$.
\texttt{Update} indicates that the slot is activated, and the slot-value is generated.
\texttt{Copy} indicates that the slot is not activated, and the slot-value is copied from the previous turn including \texttt{None}.
\texttt{Dontcare} means that the user do not care about the slot to achieve the goals.
\texttt{Delete} means that the constraint about the slot to achieve user goals is deleted, and the corresponding value reverts to \texttt{None}.
If $g_t^s = \texttt{Update}$, a GRU decoder with attention recurrently generates slot-value $V_t^s = \left\{ v_{t, 1}^s, \cdots, v_{t, \left| V_t^s \right|}^s \right\}$ for the $s$-th slot.
First, the GRU decoder generates $z_{t, j}^s$, hidden state of the $j$-th step on turn $t$ as
\begin{align}
    \label{equation10} z_{t, j}^s &= \texttt{GRU}\left( \tilde{v}_{t, j-1}^s, z_{t, j-1}^s \right) \in \mathbb{R}^{dim_H}, \quad
    \left\{ \begin{array}{l}
        \tilde{v}_{t, j-1}^s = \left\{ \begin{array}{ll}
            h_{t, s}^{Slot} \in H_t^{Slot} & \text{if } j=1 \\
            E\left( v_{t, j-1}^s \right) & \text{else}
        \end{array} \right. \\
        z_{t, 0}^s = o_t^{Init}
    \end{array} \right.,
\end{align}
where $E$ is an embedding layer.
Then, attention score $\mathbb{A}_{t, j}^s$ is calculated, and context vector $\mathbb{C}_{t, j}^s$ is obtained depending on $\mathbb{A}_{t, j}^s$ as
\begin{align}
    \label{equation11} \mathbb{A}_{t, j}^s &= \texttt{softmax}\left( H_t^{Init} z_{t, j}^s \right) \in \mathbb{R}^{\left| C_t^{Init} \right|}, \quad
    \mathbb{C}_{t, j}^s = \left( H_t^{Init} \right)^T \mathbb{A}_{t, j}^s \in \mathbb{R}^{dim_H}.
\end{align}
Finally, $v_{t, j}^s$, the $j$-th word to compose value of the $s$-th slot, is generated depending on $z_{t, j}^s$ and $\mathbb{C}_{t, j}^s$ as
\begin{align}
    \label{equation12} \tilde{z}_{t, j}^s &= W_{vocab}^{belief} \begin{bmatrix} 
        z_{t, j}^s \\
        \mathbb{C}_{t, j}^s
    \end{bmatrix} \in \mathbb{R}^{dim_V}, \quad
    v_{t, j}^s = \texttt{argmax}\left( \texttt{softmax}\left( \tilde{z}_{t, j}^s \right) \right),
\end{align}
where $W_{vocab}^{belief} \in \mathbb{R}^{dim_V \times 2dim_H}$ is an FC layer for projection to vocabulary space, and $dim_V$ is the vocabulary size.
Otherwise, $V_t^s$ is determined as follows:
\begin{align}
    \label{equation13} V_t^s = \left\{ \begin{array}{ll}
        V_{t-1}^s & \text{if } g_t^s = \texttt{Copy} \\
        \texttt{don't care} & \text{if } g_t^s = \texttt{Dontcare} \\
        \texttt{none} & \text{if } g_t^s = \texttt{Delete}
    \end{array} \right..
\end{align}
The current belief state $\mathbb{B}_t$ consists of the slot-values: $\mathbb{B}_t = \left\{ V_t^1, \cdots, V_t^{N_S} \right\}$.

\subsubsection{Dialogue Policy}
\label{subsubsec:policy}
BERT encodes $C_t^{Belief}$ to vector representations $H_t^{Belief}$, and then $h_{t, 1}^{Belief}$ that represents the \texttt{[CLS]} word is used to obtain a pooled output $o_t^{Belief}$, like Equation \ref{equation5}, as follows:
\begin{align}
    \label{equation14} H_t^{Belief} &= \texttt{BERT}\left( C_t^{Belief} \right) = \left\{ h_{t, 1}^{Belief}, \cdots, h_{t, \left| C_t^{Belief} \right| }^{Belief} \right\} \in \mathbb{R}^{\left| C_t^{Belief} \right| \times dim_H}, \\
    \label{equation15} o_t^{Belief} &= \texttt{tanh}\left( W_{pool} h_{t, 1}^{Belief} \right) \in \mathbb{R}^{dim_H}.
\end{align}
System actions $A_t = \left\{ a_{t, 1}, \cdots, a_{t, \left| A_t \right|} \right\}$ are recurrently generated by the GRU based dialogue policy with attention, like the process of belief tracker (Subsection \ref{subsubsec:belief:tracekr}), as
\begin{align}
    \label{equation16} z_{t, k} &= \texttt{GRU}\left( E\left( a_{t, k-1} \right), z_{t, k-1} \right) \in \mathbb{R}^{dim_H}, \quad \left\{ \begin{array}{l}
        a_{t, 0} = \texttt{[CLS]} \\
        z_{t, 0} = o_t^{Belief}
    \end{array} \right., \\
    \label{equation17} \mathbb{A}_{t, k} &= \texttt{softmax}\left( H_t^{Belief} z_{t, k} \right) \in \mathbb{R}^{\left| C_t^{Belief} \right|}, \quad \mathbb{C}_{t, k} = \left( H_t^{Belief} \right)^T \mathbb{A}_{t, k} \in \mathbb{R}^{dim_H}, \\
    \label{equation18} \tilde{z}_{t, k} &= W_{vocab}^{act} \begin{bmatrix}
        z_{t, k} \\
        \mathbb{C}_{t, k}
    \end{bmatrix} \in \mathbb{R}^{dim_V}, \quad
    a_{t, k} = \texttt{argmax}\left( \texttt{softmax}\left( \tilde{z}_{t, k} \right) \right),
\end{align}
where $W_{vocab}^{act} \in \mathbb{R}^{dim_V \times 2dim_H}$ is an FC layer for projection to vocabulary space.

\subsubsection{Response Generator}
\label{subsubsec:response}
BERT encodes $C_t^{Act}$ to vector representations $H_t^{Act}$, and then $h_{t, 1}^{Act}$ that represents the \texttt{[CLS]} word is used to obtain a pooled output $o_t^{Act}$, like Equation \ref{equation5}, as follows:
\begin{align}
    \label{equation19} H_t^{Act} &= \texttt{BERT}\left( C_t^{Act} \right) = \left\{ h_{t, 1}^{Act}, \cdots, h_{t, \left| C_t^{Act} \right|}^{Act} \right\} \in \mathbb{R}^{\left| C_t^{Act} \right| \times dim_H}, \\
    \label{equation20} o_t^{Act} &= \texttt{tanh}\left( W_{pool} h_{t, 1}^{Act} \right) \in \mathbb{R}^{dim_H}.
\end{align}
The response generator that uses GRU with attention recurrently generates system response $X_t = \left\{ x_{t, 1}, \cdots, x_{t, \left| X_t \right|} \right\}$ by greedy decoding, like the process of belief tracker (Subsection \ref{subsubsec:belief:tracekr}), as
\begin{align}
    \label{equation21} z_{t, l} &= \texttt{GRU}\left( E\left( x_{t, l-1} \right), z_{t, l-1} \right) \in \mathbb{R}^{dim_H}, \quad \left\{ \begin{array}{l}
        x_{t, 0} = \texttt{[CLS]} \\
        z_{t, 0} = o_t^{Act}
    \end{array} \right., \\
    \label{equation22} \mathbb{A}_{t, l} &= \texttt{softmax}\left( H_t^{Act} z_{t, l} \right) \in \mathbb{R}^{\left| C_t^{Act} \right|}, \quad \mathbb{C}_{t, l} = \left( H_t^{Act} \right)^T \mathbb{A}_{t, l} \in \mathbb{R}^{dim_H}, \\
    \label{equation23} \tilde{z}_{t, l} &= W_{vocab}^{resp} \begin{bmatrix}
        z_{t, l} \\
        \mathbb{C}_{t, l}
    \end{bmatrix} \in \mathbb{R}^{dim_V}, \quad
    x_{t, l} = \texttt{argmax}\left( \texttt{softmax}\left( \tilde{z}_{t, l} \right) \right),
\end{align}
where $W_{vocab}^{resp} \in \mathbb{R}^{dim_V \times 2dim_H}$ is an FC layer for projection to vocabulary space.

\subsection{Pre-training with SL}
\label{subsec:sl}
DORA is optimized in two steps: pre-training with SL and policy optimization with RL.
During the SL step, the entire system is trained using cross-entropy loss with gold-standard annotations.
The parameters are updated for each turn by backpropagation.
Equation \ref{equation7} is used to calculate a domain state loss $\mathcal{L}_t^{Domain}$ by using binary cross-entropy with domain state labels $Y_t^{Domain} = \left\{ y_{t, 1}^{Domain}, \cdots, y_{t, N_D}^{Domain} \right\}$ as
\begin{align}
    \label{equation24} \mathcal{L}_t^{Domain} = - \frac{1}{N_D} \sum_{d=1}^{N_D}{y_{t, d}^{Domain} logP\left( \mathbb{D}_t^d \right) + \left( 1-y_{t, d}^{Domain} \right) log\left( 1-P\left( \mathbb{D}_t^d \right) \right)},
\end{align}
where $y_{t, d}^{Domain}$ is a binary value indicating whether the $d$-th domain is activated on turn $t$.

Equation \ref{equation9} is used to calculate a slot-gate loss $\mathcal{L}_t^{Gate}$ by using cross-entropy with slot-gate labels $Y_t^{Gate} = $ \\ $\left\{ y_{t, 1}^{Gate}, \cdots, y_{t, N_S}^{Gate} \right\}$ as
\begin{align}
    \label{equation25} \mathcal{L}_t^{Gate} = - \frac{1}{N_S} \sum_{s=1}^{N_S}{\left( y_{t, s}^{Gate} \right)^T logP\left( g_t^s \right)},
\end{align}
where $y_{t, s}^{Gate} \in \mathbb{R}^4$ is an one-hot vector that indicates the $s$-th slot-gate on turn $t$.

Equation \ref{equation12} is used to calculate a slot-value loss $\mathcal{L}_t^{Value}$ by using cross-entropy with slot-value labels $Y_t^{Value}$, but $\mathcal{L}_t^{Value}$ is calculated only for \texttt{Update} slots $\mathbb{U}_t$, unlike the slot-gate loss, as
\begin{align}
    \label{equation26} \mathbb{U}_t &= \left\{ s \big\vert s \in \left[ 1, N_S \right], \quad g_t^s = \texttt{Update} \right\}, \\
    \label{equation27} Y_t^{Value} &= \left\{ y_{t, u}^{Value} \Big\vert u \in \mathbb{U}_t, \quad y_{t, u}^{Value} = \left\{ y_{t, u, 1}^{Value}, \cdots, y_{t, u, \left| V_t^u \right|}^{Value} \right\} \right\}, \\
    \label{equation28} \mathcal{L}_t^{Value} &= - \frac{1}{\left| \mathbb{U}_t \right|} \sum_{u \in \mathbb{U}_t} \frac{1}{\left| V_t^u \right|} \sum_{j=1}^{\left| V_t^u \right|} \left( y_{t, u, j}^{Value} \right)^T logP\left( v_{t, j}^u \right),
\end{align}
where $y_{t, u, j}^{Value} \in \mathbb{R}^{dim_V}$ is an one-hot vector that indicates the $j$-th word of the slot-value for the $u$-th \texttt{Update} slot on turn $t$.

The dialogue policy is trained as a conditional language model during the SL step.
Equation \ref{equation18} is used to calculate a system action loss $\mathcal{L}_t^{Act}$ by using cross-entropy with system action labels $Y_t^{Act} = \left\{ y_{t, 1}^{Act}, \cdots, y_{t, \left| A_t \right|}^{Act} \right\}$ as
\begin{align}
    \label{equation29} \mathcal{L}_t^{Act} = - \frac{1}{\left| A_t \right|} \sum_{k=1}^{\left| A_t \right|} \left( y_{t, k}^{Act} \right)^T logP\left( a_{t, k} \right),
\end{align}
where $y_{t, k}^{Act} \in \mathbb{R}^{dim_V}$ is an one-hot vector that indicates the $k$-th word of the system action on turn $t$.

Equation \ref{equation23} is used to calculate a system response loss $\mathcal{L}_t^{Resp}$ by using cross-entropy with system response labels $Y_{t}^{Resp} = \left\{ y_{t, 1}^{Resp}, \cdots, y_{t, \left| X_t \right|}^{Resp} \right\}$ as
\begin{align}
    \label{equation30} \mathcal{L}_t^{Resp} = - \frac{1}{\left| X_t \right|} \sum_{l=1}^{\left| X_t \right|} \left( y_{t, l}^{Resp} \right)^T logP\left( x_{t, l} \right),
\end{align}
where $y_{t, l}^{Resp} \in \mathbb{R}^{dim_V}$ is an one-hot vector that indicates the $l$-th word of the system response on turn $t$.

The final joint loss for pre-training with SL is sum of the above losses:
\begin{align}
    \label{equation31} \mathcal{L}_t^{SL} = \mathcal{L}_t^{Domain} + \mathcal{L}_t^{Gate} + \mathcal{L}_t^{Value} + \mathcal{L}_t^{Act} + \mathcal{L}_t^{Resp}.
\end{align}
During the SL step, DORA is trained by minimizing $\mathcal{L}_t^{SL}$ using Adam optimizer \citep{DBLP:journals/corr/KingmaB14} for each turn.

\subsection{Policy Optimization with RL}
\label{subsec:rl}
In this subsection, we describe how to optimize the pre-trained dialogue policy and to shape rewards for RL.

\subsubsection{Recurrent Action Policy Optimization}
\label{subsubsec:policy:optimization}
During the RL step, only the dialogue policy is optimized, and other modules are fixed.
The parameters are updated for each episode during the RL step because the task success is determined at the end of the conversation, whereas the parameters are updated for each turn during the SL step.
We apply REINFORCE algorithm \citep{williams1992simple}, a basic policy gradient method, to optimize the policy.
We treat the dialogue policy as a word-level policy sampling actions from a probability distribution, rather than a greedy decoder, by rewriting Equation \ref{equation18} as
\begin{align}
    \label{equation32} P_\theta\left( a_{t, k} \Big\vert C_{t}^{Belief}, \:\: a_{t, <k} \right) = \texttt{softmax}\left( \tilde{z}_{t, k} \right).
\end{align}
The policy samples a word from the probability distribution $P_\theta$, instead of using \texttt{argmax} function to select a word, to apply RL.
Even though the policy is word-level, the action space has low variance because the system actions consist of a few specific words, including some special words.
Depending on $C_t^{Belief}$, the dialogue policy recurrently samples actions until \texttt{[EOS]}, a special word to terminate the generation, is detected.
Several words comprise a system action, and several system actions comprise $A_t$ on turn $t$.
We optimize the policy to maximize rewards on collected training datasets, in other words, on the offline environment.
The policy gradient is:
\begin{align}
    \label{equation33} \nabla_\theta J(\theta) = \mathbb{E}_\theta \left[ \sum_{t=1}^T \sum_{k=1}^{\left| A_t \right|} R_{t, k} \nabla_\theta logP_\theta\left( a_{t, k} \Big\vert C_t^{Belief}, \:\: a_{t, <k} \right)\right],
\end{align}
where $T$ is the number of conversation turns in the dialogue episode, and $R_{t, k}$ is a return of $k$-th action on turn $t$.
We demonstrate the details of $R_{t, k}$ in Subsection \ref{subsubsec:reward}.
During the RL step, the policy is optimized by Equation \ref{equation33} using stochastic gradient descent (SGD) optimization algorithm.

\begin{algorithm}
\KwResult{System action rate rewards}
initialize the set of accumulated system action rates: $\Phi_A = \{\:\}$ \;
\While{RL step is not done}{
    initialize training dataset \;
    \For{each episode in training dataset}{
        \For{each turn $t$}{
            calculate system action rate $\mathcal{A}_t \in [0, 1]$ depending on $A_t$ \;
            append $\mathcal{A}_t$ to $\Phi_A$ \;
            standardize $\mathcal{A}_t$: $\mathcal{A}_t \leftarrow \frac{\mathcal{A}_t \:-\: \texttt{mean}(\Phi_A)}{\texttt{max}(\texttt{std}(\Phi_A), \:\epsilon)}$, where $\epsilon$ is a hyperparmeter \;
            assign $\mathcal{A}_t$ to system action rate reward of the last word of $A_t$: $r_{t, \left| A_t \right|}^{action} \leftarrow \mathcal{A}_t$ \;
            assign zeros to system action rate reward of the other words of $A_t$: $r_{t, 1}^{action}, \cdots, r_{t, \left| A_t \right| - 1}^{action} \leftarrow 0$ \;
        }
    }
}
\caption{Calculation of the system action rate reward}
\label{algorithm1}
\end{algorithm}

\subsubsection{Reward Shaping}
\label{subsubsec:reward}
To apply RL, we use the success rate of dialogues as a reward.
The success rate is a rate of user goals that were successfully achieved by the dialogue system, and the success rate generally indicates the main evaluation metric of task-oriented dialogue systems.
By using RL, we can directly maximize the success rate, which is non-differentiable and therefore cannot be directly set as an objective during the SL step.
However, the success rate is significantly sparse because an episode should finish to calculate it, and as a result, the RL step is inefficient and unstable.

In this work, we use the system action rate as an auxiliary reward to exploit use of explicit system actions and simultaneously to solve inefficiency and instability of the RL step.
The system action rate means the precision of generated system actions, not word-by-word, but action-by-action, and it is obtained for each turn.
Furthermore, some system actions are more important than others for achieving user goals, so for calculation of the system action rate, the system actions can be weighted according to their importance.
The total reward of the $k$-th word on turn $t$ is calculated as the weighted sum of the success rate reward $r_{t, k}^{success}$ and the system action rate reward $r_{t, k}^{action}$ as
\begin{align}
    \label{equation34} r_{t, k}^{total} = r_{t, k}^{success} + \beta r_{t, k}^{action},
\end{align}
where $\beta \in [0, 1]$ is a hyperparameter.

Rewards should be discounted over time steps to prevent bias toward long actions.
Previous methods that use a latent policy with RL discount the success rate reward over turns; i.e., the episode-level rewards calculated by the success rate assign a low weight to the first turn's actions and a high weight to the the last turn's actions.
However, this method is not clear to optimize a task-oriented dialogue system that achieves user goals across multiple turns in multi-domain dialogues.
In DORA, generated system actions do not affect the next turn's actions in offline environments because DORA does not use system responses as inputs.
Thus, we can convert the success rate reward from episode-level to turn-level by applying the same success rate reward every turn and by discounting the total reward over words for each turn as
\begin{align}
    \label{equation35} r_{1, \left| A_1 \right|}^{success} = \cdots = r_{T, \left| A_T \right|}^{success}, \quad R_{t, k} = \sum_{i=0}^{\left| A_t \right| - k} \gamma^i r_{t, k+i}^{total},
\end{align}
where $\gamma \in [0, 1]$ is a discount factor.
By using turn-level rewards, we can weight the actions for every turn equally.
The system action rate is calculated for each turn.
Algorithm \ref{algorithm1} demonstrates how to calculate $r_{t, k}^{action}$.
Unlike the system action rate, the success rate is calculated for each episode.
Algorithm \ref{algorithm2} demonstrates how to calculate $r_{t, k}^{success}$.

\begin{algorithm}
\KwResult{Success rate rewards}
initialize the set of accumulated success action rates: $\Phi_S = \{\:\}$ \;
\While{RL step is not done}{
    initialize training dataset \;
    \For{each episode in training dataset}{
        calculate success rate $\mathcal{S} \in [0, 1]$ depending on $X_1, \cdots, X_T$ \;
        append $\mathcal{S}$ to $\Phi_S$ \;
        standardize $\mathcal{S}$: $\mathcal{S} \leftarrow \frac{\mathcal{S} \:-\: \texttt{mean}(\Phi_S)}{\texttt{max}(\texttt{std}(\Phi_S), \:\epsilon)}$ \;
        assign $\mathcal{S}$ to success rate reward of the last word of $A_t$ for each turn: $r_{1, \left| A_1 \right|}^{success}, \cdots, r_{T, \left| A_T \right|}^{success} \leftarrow \mathcal{S}$ \;
        assign zeros to success rate reward of the other words of $A_t$ for each turn: $r_{1, 1}^{success}, \cdots, r_{1, \left| A_1 \right| - 1}^{success}, \cdots, r_{T, 1}^{success}, \cdots, r_{T, \left| A_T \right| - 1}^{success} \leftarrow 0$ \;
    }
}
\caption{Calculation of the success rate reward}
\label{algorithm2} 
\end{algorithm}

\subsection{System Action Control}
\label{subsec:action:control}
Explicit system actions are obviously interpretable, so they can be controlled manually.
In multi-domain task-oriented dialogues, various system actions can match a given context, and the importance of each system action can vary depending on tasks.
During inference, $A_t$ is generated by the optimized dialogue policy, and then we can modify $A_t$ depending on some rules manually designed for specific purposes.
We define simple rules by using heuristics (Subsection \ref{subsubsec:action:control}) to improve the task success on MultiWOZ datasets.

\section{Experiments}
\label{sec:experiments}
In this section, we demonstrate our experiments including datasets, evaluation metrics, detail processes, results, and ablation study.

\subsection{Experimental Setups}
\label{subsec:experimental:setups}
We used MultiWOZ 2.0 and MultiWOZ 2.1 \footnote{\url{https://github.com/budzianowski/multiwoz}.} for datasets of experiments.
MultiWOZ 2.0 is a large dataset for task-oriented dialogue systems, and MultiWOZ 2.1 is an improved version in which some annotation errors have been corrected.
MultiWOZ consists of roughly 10400 conversations that had been collected using Wizard-Of-Oz setup.
The conversations consider seven domains about travel in Cambridge, e.g., restaurants and hotels.
The dialogues in MultiWOZ are divided into about 8400 for training, 1000 for validation, and 1000 for test.

The evaluation metrics on MultiWOZ are inform rate, success rate, and BLEU score.
The inform rate counts the number of times the system provides appropriate entries that satisfy the constraints of user goals.
The success rate counts the number of times the system successfully provides information requested by users, in addition to the conditions for the inform rate.
BLEU score measures the similarity between generated responses and response labels in the dataset.
However, the response labels are not the optimal ones to achieve user goals in task-oriented dialogues.
Rather, they were simply written by humans.
Thus, BLEU score is not convincing method to evaluate task-oriented dialogue systems, so we aim to increase the inform rate and especially the success rate, rather than BLEU score.

\subsection{Experimental Details}
\label{subsec:experimental:details}
In experiments, we use BERT-base-uncased \footnote{The pre-trained model is available at \url{https://github.com/huggingface/transformers}.} as the shared encoder, and one-layer GRUs for the decoders.
MultiWOZ provides three types of setting to evaluate task-oriented dialogue systems: (1) belief tracking evaluation with gold-standard system responses, (2) policy optimization evaluation with gold-standard belief state, and (3) end-to-end evaluation without any gold-standard labels for inference.
We only evaluate DORA on the end-to-end setting, which is closest to real world problems.
We optimize DORA over two steps: pre-training during the SL step and fine-tuning during the RL step.
The SL step optimizes the entire system, whereas the RL step only optimizes the dialogue policy with other modules fixed.
We apply an early-stopping method that stops the training when validation score  does not improve over five epochs.
We conducted the experiments on a TitanRTX GPU, and the average training time was about two days for each step.

\begin{figure}
    \centering
	\includegraphics[width=0.9\textwidth]{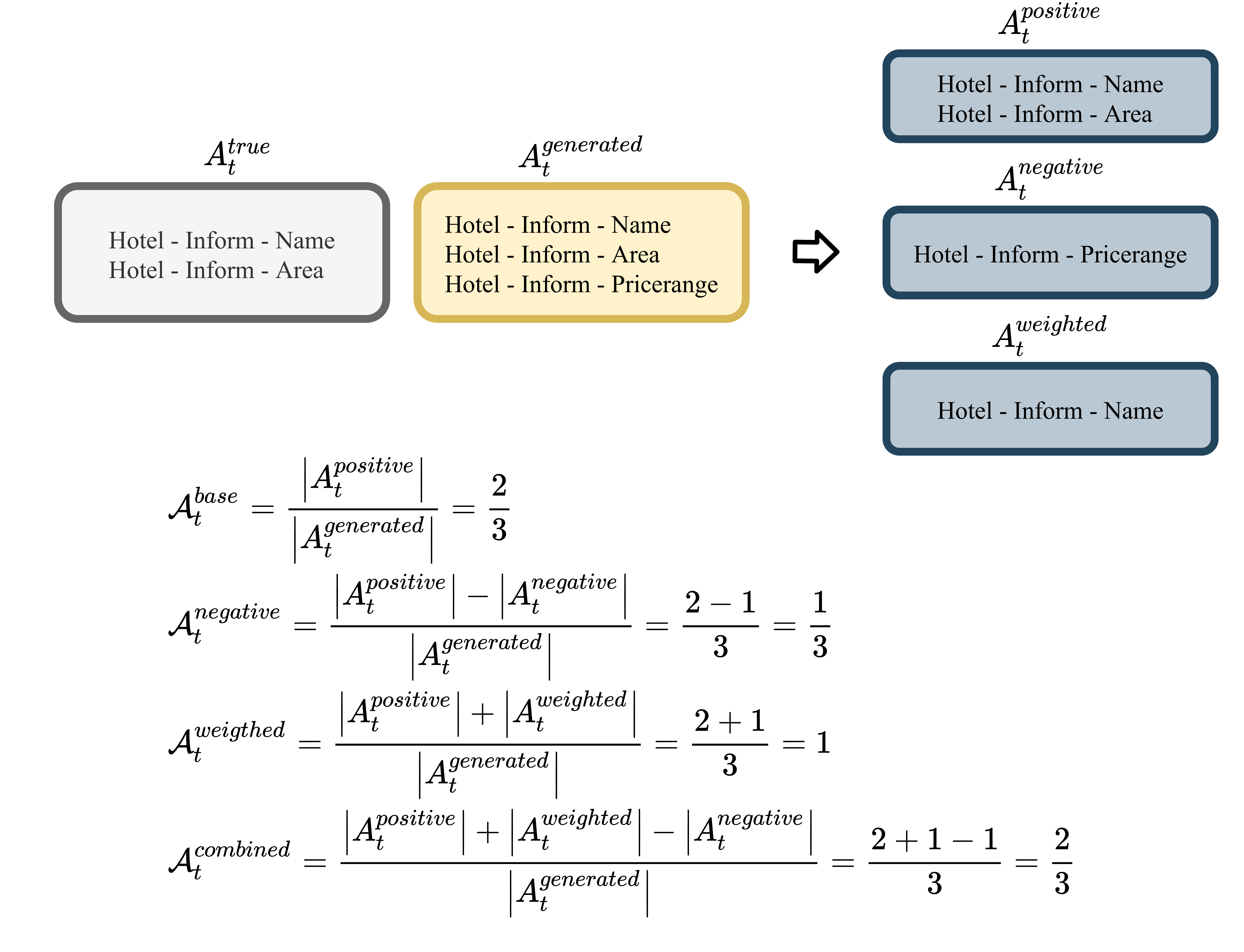}
    \caption{Process of system action rate calculation for each method on MultiWOZ.}
    \label{figure5}
\end{figure}

\subsubsection{System Action Rate}
\label{subsubsec:action:rate}
We attempted four approaches to calculating system action rate $\mathcal{A}_t$ on turn $t$.
First, the base method is to calculate precision by considering generated system actions $A_t^{generated}$ and system action labels $A_t^{true}$ as follows:
\begin{align}
    \label{equation36} A_t^{positive} = \left\{ a \big\vert a \in A_t^{generated} \cap A_t^{true} \right\}, \quad \mathcal{A}_t^{base} = \frac{\left| A_t^{positive} \right|}{\left| A_t^{generated} \right|}.
\end{align}
Second, the negative method gives a penalty for system actions that do not exist in $A_t^{true}$ to prevent the dialogue policy from generating too many system actions as follows:
\begin{align}
    \label{equation37} A_t^{negative} = \left\{ a \big\vert a \in A_t^{generated} - A_t^{true} \right\}, \quad \mathcal{A}_t^{negative} = \frac{\left| A_t^{positive} \right| - \left| A_t^{negative} \right|}{\left| A_t^{generated} \right|}.
\end{align}
Third, the weighted method increases the weight given to system actions that are important to achieve task success as:
\begin{align}
    \label{equation38} A_t^{weighted} = \left\{ a \big\vert a \in A_t^{positive} \cap A^{important} \right\}, \quad \mathcal{A}_t^{weighted} = \frac{\left| A_t^{positive} \right| + \left| A_t^{weighted} \right|}{\left| A_t^{generated} \right|},
\end{align}
where the set of important system actions $A^{important}$ is predefined by heuristic (Tabel \ref{table7}).
Finally, the combined negative and weighted method combines the two methods as follows:
\begin{align}
    \label{equation39} \mathcal{A}_t^{combined} = \frac{\left| A_t^{positive} \right| + \left| A_t^{weighted} \right| - \left| A_t^{negative} \right|}{\left| A_t^{generated} \right|}.
\end{align}
Figure \ref{figure5} shows examples of system action rate calculation process.
Each system action is a triple of domain, action, and slot.
In Figure \ref{figure5}, three system actions are generated by the dialogue policy: \texttt{Hotel-Inform-Name}, \texttt{Hotel-Inform-Area}, and \texttt{Hotel-Inform-Pricerange}.

\begin{figure}
    \centering
	\includegraphics[width=0.9\textwidth]{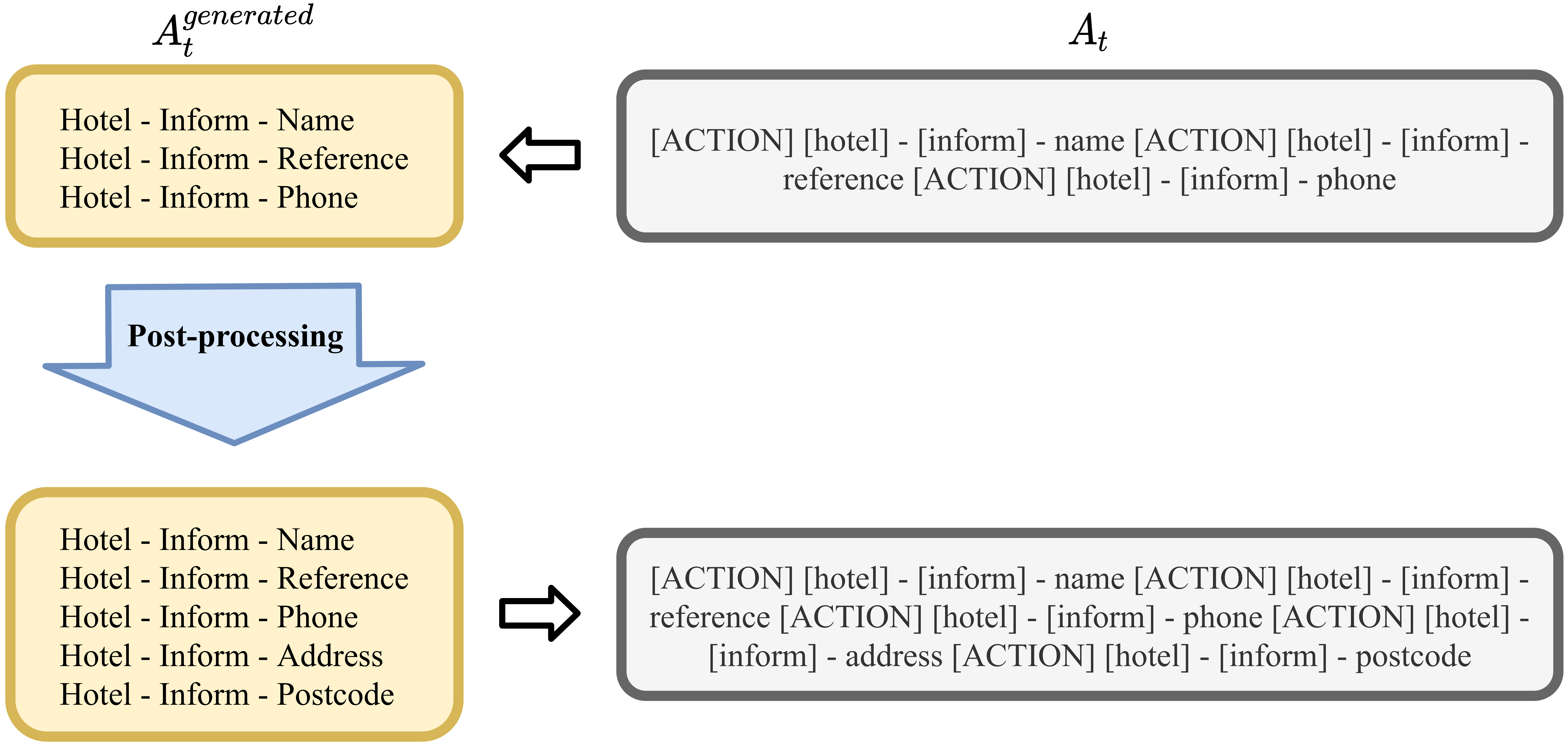}
    \caption{Process of system action control by post-processing on MultiWOZ.}
    \label{figure6}
\end{figure}

\subsubsection{System Action Control}
\label{subsubsec:action:control}
We control generated system actions by post-processing depending on some rules.
By this control of system actions, we aim to improve the success rate, which is the main evaluation metric on MultiWOZ and is matched with the objective of task-oriented dialogue systems in the real world.
Several system actions are more important than others for the task success on MultiWOZ; e.g., to provide a phone number requested by the user.
In our experiments, we define simple post-processing rules that adjust generated system actions:
\begin{enumerate}
    \item If the dialogue policy generates system actions to provide some information about a hotel,
    \item Adjust the actions to necessarily include phone number, address, and postcode of the hotel.
    \item Repeat 1-2 for restaurants and attractions.
\end{enumerate}
Figure \ref{figure6} shows how $A_t$ is parsed to $A_t^{generated}$ and how the generated system actions are controlled by post-processing.
We apply the post-processing only during inference, not training.
In this paper, we only attempt to simply control system actions for the task success.

\begin{table}[ht]
    \renewcommand{\tabcolsep}{5pt}
    \centering
    \caption{Comparison of various configurations of DORA on test set of MultiWOZ 2.0 and MultiWOZ 2.1.}
    \begin{tabular}{c|l|ccc|ccc}
        \Xhline{5\arrayrulewidth}
        \hline \multicolumn{2}{c|}{} & \multicolumn{3}{c|}{MultiWOZ 2.0} & \multicolumn{3}{c}{MultiWOZ 2.1} \\
        \cline{3-8} \multicolumn{2}{c|}{} & Inform$\uparrow$ & Success$\uparrow$ & BLEU$\uparrow$ & Inform$\uparrow$ & Success$\uparrow$ & BLEU$\uparrow$ \\
        \hline \multirow{5}{*}{w/o post-processing} & w/o action rate & 94.9 & 90.0 & 12.61 & 94.8 & 90.4 & 13.29 \\
        \cline{2-8} & base & 94.6 & \textbf{92.0} & 12.70 & 94.4 & 91.1 & 12.58 \\
        \cline{2-8} & negative & 94.1 & 88.9 & \textbf{13.94} & 94.3 & 90.4 & \textbf{13.40} \\
        \cline{2-8} & weighted & 94.9 & 91.4 & 12.23 & \textbf{95.2} & 92.3 & 12.31 \\
        \cline{2-8} & combined & \textbf{95.3} & 91.3 & 13.31 & 95.0 & 91.4 & 12.80 \\
        \hline\hline \multirow{5}{*}{w/ post-processing} & w/o action rate & 94.9 & 91.9 & 12.58 & 94.8 & 91.6 & 12.95 \\
        \cline{2-8} & base & 94.7 & \textbf{92.0} & 12.72 & 94.4 & 91.0 & 12.46 \\
        \cline{2-8} & negative & 94.0 & 90.6 & 12.78 & 94.3 & 90.8 & 13.10 \\
        \cline{2-8} & weighted & 94.9 & \textbf{92.0} & 12.17 & \textbf{95.2} & \textbf{92.7} & 12.21 \\
        \cline{2-8} & combined & 95.2 & 91.6 & 13.26 & 95.0 & 91.9 & 12.20 \\
        \Xhline{5\arrayrulewidth}
    \end{tabular}
    \label{table1}
\end{table}

\begin{table}
    \renewcommand{\tabcolsep}{4pt}
    \centering
    \caption{Results of end-to-end evaluation on test set of MultiWOZ 2.0 and MultiWOZ 2.1. SUMBT+LaRL uses a previous system response instead of the entire history.}
    \begin{tabular}{l|c|ccc|ccc}
        \Xhline{5\arrayrulewidth} \multirow{2}{*}{Model} & \multirow{2}{*}{History} & \multicolumn{3}{c|}{MultiWOZ 2.0} & \multicolumn{3}{c}{MultiWOZ 2.1}\\
        \cline{3-8} & & Inform$\uparrow$ & Success$\uparrow$ & BLEU$\uparrow$ & Inform$\uparrow$ & Success$\uparrow$ & BLEU$\uparrow$ \\
        \hline DAMD \citep{zhang2020task} & \checkmark & 76.30 & 60.40 & 16.60 & - & - & - \\
        LABES-S2S \citep{zhang2020probabilistic} & \checkmark & - & - & - & 78.07 & 67.06 & \textbf{18.13} \\
        SimpleTOD \citep{hosseini2020simple} & \checkmark & 84.40 & 70.10 & 15.01 & 85.00 & 70.50 & 15.23 \\
        SOLOIST \citep{peng2020soloist} & \checkmark & 85.50 & 72.90 & 16.54 & - & - & - \\
        MinTL-BART \citep{lin2020mintl} & \checkmark & 84.88 & 74.91 & 17.89 & - & - & - \\
        LAVA \citep{lubis2020lava} & \checkmark & 91.80 & 81.80 & 12.03 & - & - & - \\
        UBAR \citep{yang2020ubar} & \checkmark & \textbf{95.40} & 80.70 & 17.00 & \textbf{95.70} & 81.80 & 16.50 \\
        SUMBT+LaRL \citep{lee2020sumbt+} & \checkmark & 92.20 & 85.40 & \textbf{17.90} & - & - & - \\
        \hline DORA & & 94.60 & \textbf{92.00} & 12.70 & 94.40 & \textbf{91.10} & 12.58 \\
        \Xhline{5\arrayrulewidth}
    \end{tabular}
    \label{table2}
\end{table}

\subsection{Experimental Results}
\label{subsec:experimental:results}
Table \ref{table1} compares the results of four methods for calculating the system action rate as mentioned in Subsection \ref{subsubsec:action:rate} and the results without the system action rate as reward for the RL step.
The results further contains the effects of system action control by post-processing.
In our experiments, use of the system action rate as a reward for RL quite improved the success rate except of the negative method, but it had little effect on the success rate when we applied system action control in addition to the use of the system action rate.
These results suggest that use of the system action rate and control of system actions had similar effects in the experiments by increasing the weights of certain system actions to enable the task success.
Also, we believe that the negative method gave wrong penalties for system actions that did not exist in true system actions of datasets even though they were appropriate system actions.

In our experiments, DORA achieved higher task success than previous methods on end-to-end evaluation setting of MultiWOZ 2.0 and MultiWOZ 2.1 without any previous utterances (Table \ref{table2}).
DORA also achieved higher task success during the SL step than previous methods using latent actions for RL even though several of the previous methods used gold-standard belief state for inference (Table \ref{table3}).
These results demonstrate that pre-training with latent policy is unstable, whereas DORA becomes well optimized during the SL step.

\begin{table}
    \renewcommand{\tabcolsep}{10pt}
    \centering
    \caption{SL step results of previous models with latent actions and DORA on MultiWOZ 2.0.}
    \begin{tabular}{l|c|ccc}
        \Xhline{5\arrayrulewidth} Model & Belief State & Inform$\uparrow$ & Success$\uparrow$ & BLEU$\uparrow$ \\
        \hline LaRL & oracle & 67.98 & 57.36 & 19.10 \\
        LAVA & oracle & 71.97 & 57.96 & 18.00 \\
        SUMBT+LaRL & generated & 72.10 & 66.20 & \textbf{19.36} \\
        \hline DORA & generated & \textbf{85.60} & \textbf{74.60} & 15.35 \\
        \Xhline{5\arrayrulewidth}
    \end{tabular}
    \label{table3}
\end{table}

\begin{figure}
    \centering
    \subfloat[Lengths of two system inputs on MultiWOZ.]{
    \label{subfig7a}
	\includegraphics[width=0.48\textwidth]{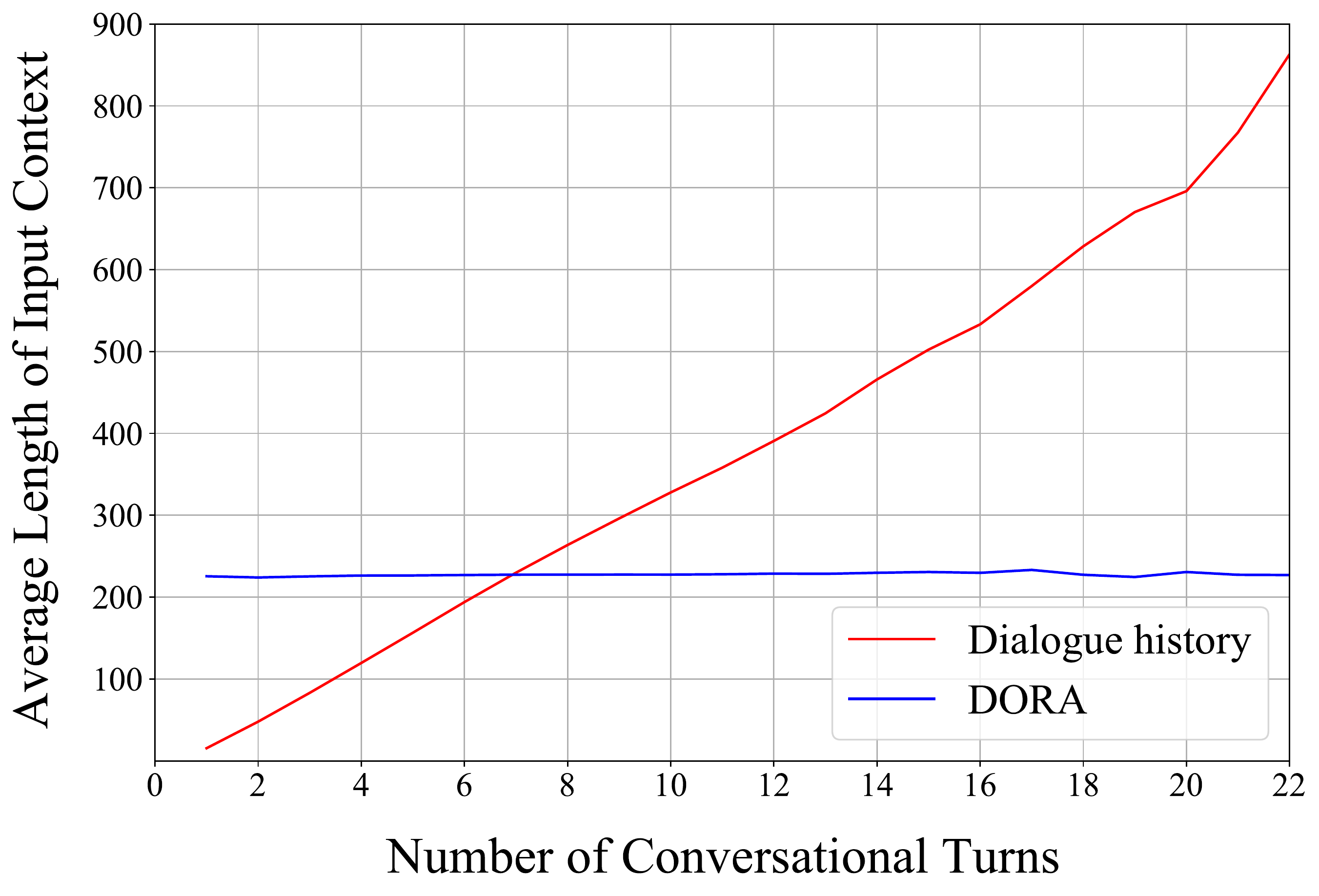}}
	\subfloat[Memory usage depending on input length.]{
    \label{subfig7b}
	\includegraphics[width=0.49\textwidth]{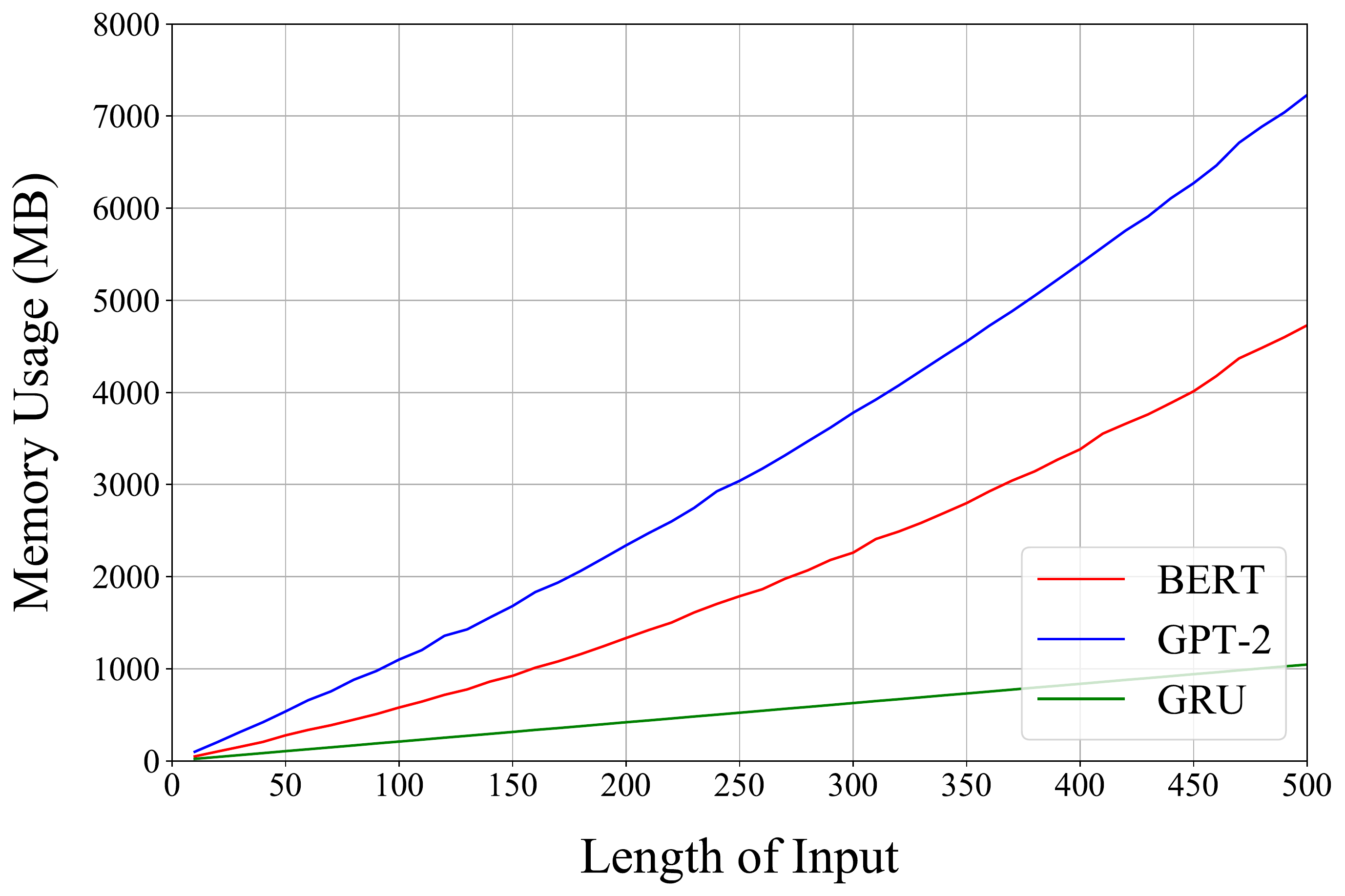}}
	\caption{Input length comparison and memory usage acceleration depending on input length.}
    \label{figure7}
\end{figure}

We conducted additional experiments to confirm the efficiency of our input context.
As the number of turns increased, the length of input context used in DORA was almost constant, whereas the dialogue history lengthened (Figure \ref{subfig7a}).
We compared memory usage depending on the input length on three language models: BERT, GPT-2, and GRU (Figure \ref{subfig7b}).
The memory usage of the three language models increased with different speeds as the input length increased.
The language models have 12 layers with hidden size of 768, but GPT-2 uses a larger vocabulary than the other two.
The memory usage indicates the allocated amount of memory on GPU, except to retain the model, after an input with batch size of 8 is fed into the model.
Even with same increment of input length, the memory usage increased strongly as model was large.

\subsubsection{Ablation Study}
\label{subsubsec:ablation}
To verify the effectiveness of our approach, we conducted an ablation study (Table \ref{table4}).
We sequentially removed system action rate, RL step, and the domain state.
Furthermore, we attempted using the dialogue history instead of our efficient context, as in previous methods.
Removal of the system action rate from rewards for the RL step slightly decreased the success rate.
Subsequent removal of the RL step and then optimization using only the SL greatly degraded the success rate.
Additionally, removal of the domain state from the input context further decreased the success rate.
Finally, when dialogue history was used as input, the success rate was lower than when we used the efficient context.
The results show that the components of our approach help to improve the success rate.

\begin{table}[ht]
    \renewcommand{\tabcolsep}{10pt}
    \centering
    \caption{Ablation study on MultiWOZ 2.0.}
    \begin{tabular}{l|ccc}
        \Xhline{5\arrayrulewidth} & Inform$\uparrow$ & Success$\uparrow$ & BLEU$\uparrow$ \\
        \hline DORA & 94.6 & 92.0 & 12.70 \\
        - system action rate & 94.9 & 90.0 & 12.61 \\
        - RL & 85.6 & 74.6 & 15.35 \\
        - domain state & 85.2 & 68.5 & 14.14 \\
        using history & 79.3 & 63.5 & 14.48 \\
        \Xhline{5\arrayrulewidth}
    \end{tabular}
    \label{table4}
\end{table}

\section{Discussion}
The SL step is the cornerstone of the RL step, and explicit system action policy makes the SL step clearer (Table \ref{table3}).
The system action policy further enables use of system actions to shape rewards during the RL step and to control the system actions generated by the optimized policy.
Use of the system action rate as reward generally improved success rate except when the negative method was applied (Table \ref{table1}).
Calculation of the system action rate used labels that had been annotated on datasets as $A_t^{true}$.
On turn $t$, $A_t^{true}$ is a set of appropriate system actions, but it is not the optimal one.
This problem seems to be similar to the bias that occurs when SL is used to train task-oriented dialogue systems.
The negative method may have given wrong penalties for system actions that were appropriate on turn $t$ but not included in the labels.

In our experiments, system action control improved the success rate more when rewards were calculated without the system action rate, than with it.
The purpose of system action control was to improve the task success on MultiWOZ (Subsection \ref{subsubsec:action:control}).
Use of the system action rate as a reward for RL has a semantically similar purpose: to give more rewards when the dialogue policy generates appropriate system actions for the task success.
Therefore, system action control improved the success rate more when reward was shaped without the system action rate, and use of the system action rate improved the success rate more when system action was not controlled (Table \ref{table1}).
The results show that use of the system action rate and control of system actions performed semantically similar roles for task success; i.e., the system action control fulfilled our purpose even though it depended on a simple heuristic.

Furthermore, control of the optimized dialogue policy by post-processing enables the system to operate as a hybrid system.
Recently, deep learning methods have begun to be more successful than conventional rule-based methods.
Models that use deep learning operate well in general and cover wide domains.
However, the models are not perfect and have not completely replaced rule-based models in the real world.
The conventional models are more accurate than the deep learning models in specific areas.
Also, deep learning models require large cost for training on new areas.
Therefore, research to develop hybrid approaches that have advantages of both approaches may have significant benefits, and we believe that our approach has value as a way to solve practical problems, in addition to its increase in the success rate.

Use of pre-trained language models has been a trend in NLP field, including task-oriented dialogues.
The strategy has improved the success rate, but large models require a large amount of memory (Figure \ref{subfig7b}) and thereby increase the training cost.
Use of the dialogue history makes memory usage unstable and thereby impedes memory management on GPUs, in addition to increasing the memory usage, especially on large models.
Furthermore, using the dialogue history as input induces a dependency on generated system responses during inference, as an NLU module encodes the generated response on next turn.
This observation seems counterintuitive because the responses were generated by depending on information already known to the system.
Also, the use of dialogue history as input can cause propagation of errors to future turns.
For these reasons, approaches to efficient input context should be sought for use in task-oriented dialogue systems.

We construct the input context by combining the previous domain state, belief state, and the current user utterance.
The previous states contain abstract information accumulated during previous turns, and the current user utterance indicates new information from the user.
By using the context, the system can efficiently track the flow of conversation by comparing them and then updating the states.
Furthermore, the context stabilizes the system by removing dependency on generated system responses.

\section{Conclusion}
In this paper, we have proposed DORA, which is an efficient task-oriented dialogue system that uses effective optimization methods.
We have demonstrated that clear pre-training with SL is important for effective fine-tuning with RL, and the explicit system action policy clarifies the SL step.
Our experiments show that the task success is increased by using the system action rate to shape rewards during the RL step, and by post-processing to control system actions.
We have further presented a fresh perspective of task-oriented dialogue systems as a hybrid approach to address practical problems as well as academic research.
Also, we have proposed an efficient method to construct input context that can replace the entire dialogue history.

\section*{Declaration of Competing Interest}
The authors declare that they have no known competing financial interests or personal relationships that could have appeared to influence the work reported in this paper.

\section*{Acknowledgement}
This work was supported by Institute of Information \& communications Technology Planning \& Evaluation (IITP) grant funded by the Korea government(MSIT) (No.2019-0-01906, Artificial Intelligence Graduate School Program(POSTECH)); and the MSIT(Ministry of Science and ICT), Korea, under the ITRC(Information Technology Research Center) support program(IITP-2021-2020-0-01789) supervised by the IITP(Institute for Information \& Communications Technology Planning \& Evaluation)

\appendix

\section{Hyperparameters}
\label{sec:appendix:hyperparams}
We report the hyperparameters used in the experiments for reproducibility.
Table \ref{table5} lists the hyperparameters for the SL and RL steps.

\begin{table}[ht]
    \renewcommand{\tabcolsep}{10pt}
    \centering
    \caption{Hyperparameters used for the experiments in this paper.}
    \begin{tabular}{l|c|l}
        \Xhline{5\arrayrulewidth}
        \multicolumn{3}{c}{SL step} \\
        \hline Hidden size & 768 & The size of hidden layers. \\
        Embedding size & 768 & The size of embedding layer. \\
        Vocabulary size & 30522 & The size of vocabulary. \\
        Max context length & 512 & The maximum length of input context. \\
        Dropout & 0.2 & The rate of dropout. \\
        Early stopping count & 5 & The maximum count to early stop the training. \\
        Max epochs & 40 & The maximum number of epochs. \\
        Min epochs & 20 & The minimum number of epochs ignoring the early stopping. \\
        Optimizer & Adam & The type of optimizer for the SL step. \\
        Batch size & 8 & The size of mini batch for the SL step. \\
        Learning rate & 3e-5 & The learing rate during the SL step. \\
        Gradient clipping & 10 & The maximum norm of gradient during the SL step. \\
        \hline\hline \multicolumn{3}{c}{RL step} \\
        \hline $\beta$ (MultiWOZ 2.0) & 1e-3 & \multirow{2}{*}{The weight of action rate to calculate the total reward.} \\
        $\beta$ (MultiWOZ 2.1) & 1e-2 & \\
        $\gamma$ & 0.99 & The discount factor during the RL step. \\
        Optimizer & SGD & The type of optimizer for the RL step. \\
        Batch size & 8 & The size of mini batch for the RL step. \\
        Learning rate & 1e-2 & The learing rate during the RL step. \\
        Gradient clipping & 1 & The maximum norm of gradient during the RL step. \\
        $\epsilon$ & 1e-4 & The epsilon for standardization of rewards during the RL step. \\
        \Xhline{5\arrayrulewidth}
    \end{tabular}
    \label{table5}
\end{table}

\section{Input Context Flow}
\label{sec:appendix:context}
\begin{figure}[ht]
    \centering
	\includegraphics[width=0.9\textwidth]{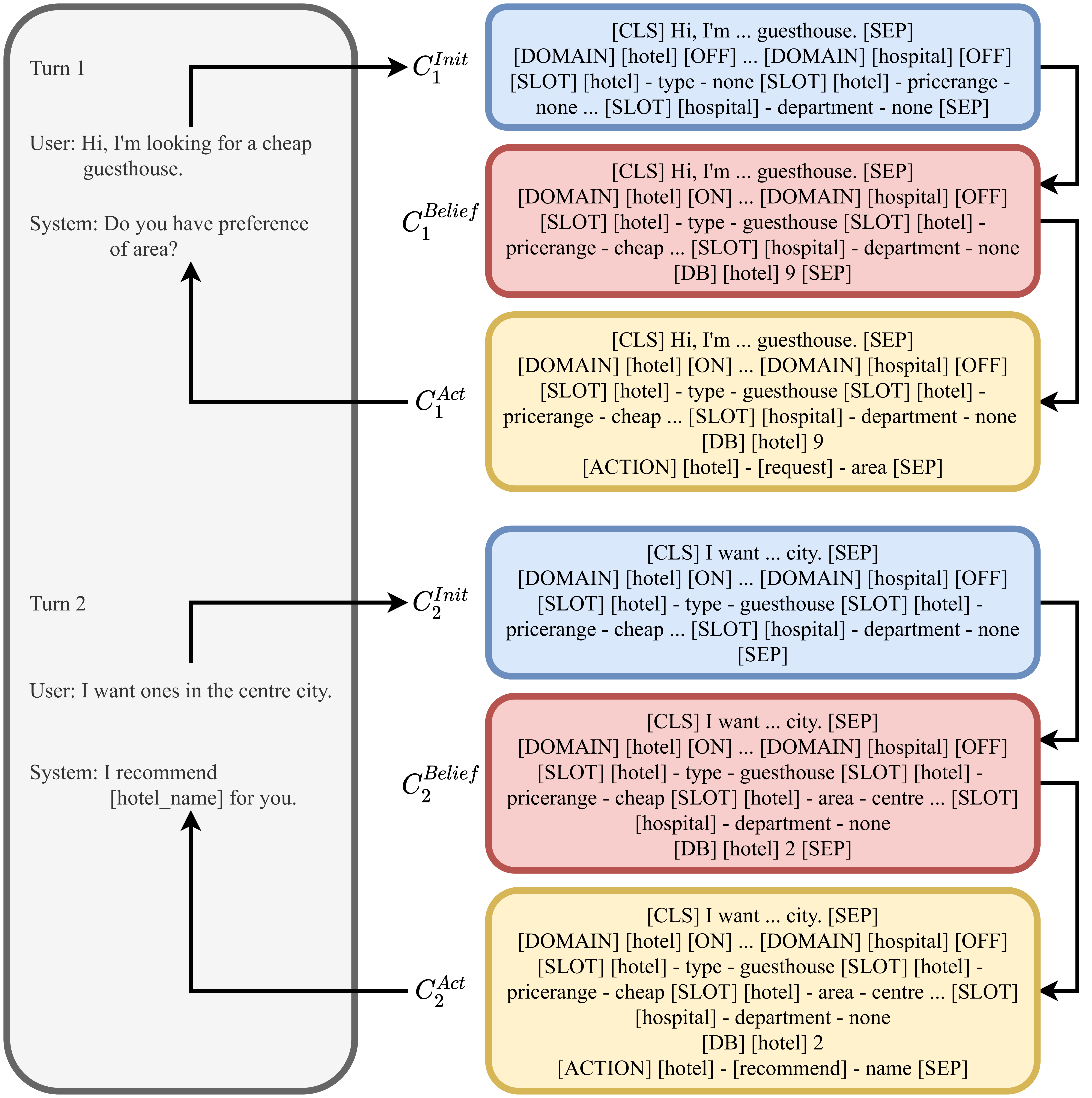}
    \caption{Flow of the context indicating the input sequence of shared BERT encoder on MultiWOZ. The words enclosed in brackets are special words.}
    \label{figure8}
\end{figure}

Figure \ref{figure8} shows an example of flow of the input context: initial context $C_t^{Init}$, belief context $C_t^{Belief}$, and action context $C_t^{Act}$.
On the initial turn, each domain in the previous domain state $\mathbb{D}_0$ is OFF, and each slot-value in the previous belief state $\mathbb{B}_0$ is \texttt{None}.

\newpage

\section{Definition of System Action Set}
\label{sec:appendix:action:set}
We report the set of system actions defined on MultiWOZ.
Tabel \ref{table6} lists the system actions annotated on MultiWOZ for each domain.
We define the set of important system actions $A^{important}$ to calculate the weighted system action rate.
Table \ref{table7} lists the system actions in $A^{important}$ regarded as important actions in terms of the task success on MultiWOZ.

\begin{longtable}{c|c|cccccc}
    \caption{Set of system actions defined on MultiWOZ.}
    \label{table6} \\
        \hlineB{5}
        Domain & Action & \multicolumn{6}{c}{Slot} \\
        \hline\hline \multirow{2}{*}{Hotel} & \multirow{2}{*}{Inform} & Name & Type & Area & Pricerange & Internet & Parking \\
        & & Address & Postcode & Phone & Stars & Choice & Reference \\ \cline{2-8}
        \newpage
        \cline{2-8} \multirow{8}{*}{Hotel} & \multirow{2}{*}{Request} & Name & Type & Area & Pricerange & Internet & Parking \\
        & & Start \\
        
        \cline{2-8} & \multirow{2}{*}{Recommend} & Name & Type & Area & Pricerange & Internet & Parking \\
        & & Address & Postcode & Phone & Stars & Choice \\
        \cline{2-8} & \multirow{2}{*}{Select} & Name & Type & Area & Pricerange & Internet & Parking \\
        & & Address & Phone & Stars & Choice \\
        \cline{2-8} & \multirow{2}{*}{NoOffer} & Name & Type & Area & Pricerange & Internet & Parking \\
        & & Stars & Choice \\
        
        \hline \multirow{7}{*}{Restaurant} & \multirow{2}{*}{Inform} & Name & Food & Area & Pricerange & Address & Postcode \\
        & & Phone & Choice & Reference \\
        \cline{2-8} & Request & Name & Food & Area & Pricerange \\
        \cline{2-8} & \multirow{2}{*}{Recommend} & Name & Food & Area & Pricerange & Address & Postcode \\
        & & Phone & Choice \\
        \cline{2-8} & Select & Name & Food & Area & Pricerange & Address & Choice \\
        \cline{2-8} & NoOffer & Name & Food & Area & Pricerange & Choice \\
    
        \hline \multirow{8}{*}{Attraction} & \multirow{2}{*}{Inform} & Name & Type & Area & Price & Open & Address \\
        & & Postcode & Phone & Choice \\
        \cline{2-8} & Request & Name & Type & Area & Price \\
        \cline{2-8} & \multirow{2}{*}{Recommend} & Name & Type & Area & Price & Open & Address \\
        & & Postcode & Phone & Choice \\
        \cline{2-8} & \multirow{2}{*}{Select} & Name & Type & Area & Price & Address & Phone \\
        & & Choice \\
        \cline{2-8} & NoOffer & Name & Type & Area & Price & Address & Choice \\
        
        \hline \multirow{11}{*}{Train} & \multirow{2}{*}{Inform} & Id & ArriveBy & LeaveAt & Departure & Destination & Duration \\
        & & Day & People & Price & Choice & Reference \\
        \cline{2-8} & Request & ArriveBy & LeaveAt & Departure & Destination & Day & People \\
        \cline{2-8} & \multirow{2}{*}{OfferBooked} & Id & ArriveBy & LeaveAt & Departure & Destination & Duration \\
        & & Day & People & Price & Choice & Reference \\
        \cline{2-8} & \multirow{2}{*}{OfferBook} & Id & ArriveBy & LeaveAt & Departure & Destination & Duration \\
        & & Day & People & Price & Choice & Reference \\
        \cline{2-8} & \multirow{2}{*}{Select} & Id & ArriveBy & LeaveAt & Departure & Destination & Day \\
        & & People & Price & Choice \\
        \cline{2-8} & \multirow{2}{*}{NoOffer} & Id & ArriveBy & LeaveAt & Departure & Destination & Day \\
        & & Choice \\
        
        \hline \multirow{2}{*}{Taxi} & Inform & ArriveBy & LeaveAt & Departure & Destination & Car & Phone \\
        \cline{2-8} & Request & ArriveBy & LeaveAt & Departure & Destination \\
        
        \hline \multirow{2}{*}{Hospital} & Inform & Department & Address & Postcode & Phone \\
        \cline{2-8} & Request & Department \\
        
        \hline Police & Inform & Name & Address & Postcode & Phone \\
        
        \hline \multirow{4}{*}{Booking} & Inform & Name & Time & Stay & Day & People & Reference \\
        \cline{2-8} & Request & Time & Stay & Day & People \\
        \cline{2-8} & Book & Name & Time & Stay & Day & People & Reference \\
        \cline{2-8} & NoBook & Name & Time & Stay & Day & People & Reference \\
        
        \hline \multirow{4}{*}{General} & Bye & \multicolumn{6}{c}{-} \\
        \cline{2-8} & Greet & \multicolumn{6}{c}{-} \\
        \cline{2-8} & Reqmore & \multicolumn{6}{c}{-} \\
        \cline{2-8} & Welcome & \multicolumn{6}{c}{-} \\
        \hlineB{5}
\end{longtable}

\begin{table}
    \renewcommand{\tabcolsep}{10pt}
    \centering
    \caption{Predefined important system actions on MultiWOZ.}
    \begin{tabular}{c|c|ccccc}
        \Xhline{5\arrayrulewidth}
        Domain & Action & \multicolumn{5}{c}{Slot} \\
        \hline\hline Hotel & Inform & Name & Address & Postcode & Phone & Reference \\
        \hline Restaurant & Inform & Name & Address & Postcode & Phone & Reference \\
        \hline Attraction & Inform & Name & Address & Postcode& Phone & \\
        \hline \multirow{3}{*}{Train} & Inform & Id & Reference \\
        \cline{2-7} & OfferBooked & Id & Reference \\
        \cline{2-7} & OfferBook & Id \\
        \hline Taxi & Inform & Phone \\
        \hline \multirow{2}{*}{Booking} & Inform & Name \\
        \cline{2-7} & Book & Name & Reference \\
        \Xhline{5\arrayrulewidth}
    \end{tabular}
    \label{table7}
\end{table}

\newpage

\section{Sample Dialogues}
\label{sec:appendix:sample:dialogues}
We report sample dialogues generated by DORA.
Table \ref{table8} shows the results on PMUL2275 dialogue of MultiWOZ.
Table \ref{table9}, \ref{table10} show the results on interactive dialogues with a user.

\begin{longtable}{ll}
    \caption{Sample generation results of DORA on the PMUL2275 dialogue of test set.}
    \label{table8} \\
        \hlineB{10}
        User: & what is the address for holy trinity church? \\
        \hline Belief: & \{attraction-name: holy trinity church\} \\
        \hline \multirow{2}{*}{Action:}
        & [attraction]-[inform]-postcode, [attraction]-[inform]-address, [attraction]-[inform]-name, \\ 
        & [general]-[reqmore] \\
        \hline \multirow{2}{*}{Response:}
        & [attraction\_name] is located on [attraction\_address], postcode [attraction\_postcode]. \\ 
        & do you need any other information? \\
        \hlineB{5}
        User: & thank you. i wish to get entrance fee \\
        \hline Belief: & \{attraction-name: holy trinity church\} \\
        \hline \multirow{2}{*}{Action:}
        & [attraction]-[inform]-price, [attraction]-[inform]-phone, [attraction]-[inform]-name, \\
        & [attraction]-[inform]-postcode \\
        \hline \multirow{2}{*}{Response:}
        & [attraction\_name] has free admission. their phone number is [attraction\_phone]. \\
        & the postcode is [attraction\_postcode]. \\
        \hlineB{5}
        User: & yes, i am looking for an indian restaurant in the east. \\
        \hline Belief: & \{restaurant-food: indian, restaurant-area: east, attraction-name: holy trinity church\} \\
        \hline \multirow{2}{*}{Action:}
        & [booking]-[inform], [restaurant]-[inform]-name, [restaurant]-[inform]-address, \\
        & [restaurant]-[inform]-phone \\
        \hline \multirow{2}{*}{Response:}
        & i have the [restaurant\_name] located at [restaurant\_address]. the phone number is \\ & [restaurant\_phone]. would you like to make reservations? \\
        \hlineB{5}
        User: & how about an expensive one. what are my choices in that range? \\
        \hline \multirow{2}{*}{Belief:}
        & \{restaurant-food: indian, restaurant-pricerange: expensive, restaurant-area: east, \\ 
        & attraction-name: holy trinity church\} \\
        \hline \multirow{2}{*}{Action:}
        & [booking]-[inform], [restaurant]-[inform]-name, [restaurant]-[inform]-address, \\
        & [restaurant]-[inform]-phone, [restaurant]-[inform]-phone \\
        \hline \multirow{2}{*}{Response:}
        & there is one. [restaurant\_name] is located at [restaurant\_address]. their phone \\ 
        & number is [restaurant\_phone]. would you like to book a table? \\
        \hlineB{5}
        User: & i want to book for 7 people at 18:00 on monday \\
        \hline \multirow{3}{*}{Belief:}
        & \{restaurant-time: 18:00, restaurant-day: monday, restaurant-people: 7, restaurant-food: indian, \\
        & restaurant-pricerange: expensive, restaurant-area: east, attraction-name: holy trinity church\} \\
        \hline Action: & [booking]-[book]-reference, [booking]-[book]-name \\
        \hline Response:
        & i have booked you at [restaurant\_name]. your reference number is [restaurant\_reference]. \\
        \hlineB{5}
        User:
        & yes, one more thing. i will need a taxi from holy trinity church to pipasha restaurant. \\
        \hline \multirow{3}{*}{Belief:}
        & \{taxi-destination: pipasha restaurant, taxi-departure: holy trinity church, restaurant-time: 18:00, \\
        & restaurant-day: monday, restaurant-people: 7, restaurant-food: indian, \\
        & restaurant-pricerange: expensive, restaurant-area: east, attraction-name: holy trinity church\} \\
        \hline Action: & [taxi]-[request]-arrive, [taxi]-[request]-leave \\
        \hline Response: & what time would you like to leave or arrive by? \\
        \hlineB{5}
        User: & i need to arrive by the booked time at the restaurant. \\
        \hline \multirow{3}{*}{Belief:}
        & \{taxi-destination: pipasha restaurant, taxi-departure: holy trinity church, taxi-arrive: 18:00, \\
        & restaurant-time: 18:00, restaurant-day: monday, restaurant-people: 7, restaurant-food: indian, \\
        & restaurant-pricerange: expensive, restaurant-area: east, attraction-name: holy trinity church\} \\
        \hline Action:
        & [general]-[reqmore], [taxi]-[inform]-phone, [taxi]-[inform]-car \\
        \hline \multirow{2}{*}{Response:}
        & i have booked a [taxi\_car] for you. the contact number is [taxi\_phone]. is there anything else \\
        & i can help you with? \\
        \hlineB{5}
        User: & there will be 7 people travelling. \\
        \hline \multirow{3}{*}{Belief:}
        & \{taxi-destination: pipasha restaurant, taxi-departure: holy trinity church, taxi-arrive: 18:00, \\
        & restaurant-time: 18:00, restaurant-day: monday, restaurant-people: 7, restaurant-food: indian, \\
        & restaurant-pricerange: expensive, restaurant-area: east, attraction-name: holy trinity church\} \\
        \hline Action:
        & [taxi]-[inform]-phone, [taxi]-[inform]-car \\
        \hline Response: & i have booked a [taxi\_car] for you. the contact number is [taxi\_phone]. \\
        \hlineB{5}
        User: & thank you. that is all i need. \\
        \hline \multirow{3}{*}{Belief:}
        & \{taxi-destination: pipasha restaurant, taxi-departure: holy trinity church, taxi-arrive: 18:00, \\
        & restaurant-time: 18:00, restaurant-day: monday, restaurant-people: 7, restaurant-food: indian, \\
        & restaurant-pricerange: expensive, restaurant-area: east, attraction-name: holy trinity church\} \\
        \hline Response: & you're welcome, have a great day! \\
        \hlineB{10}
\end{longtable}

\begin{longtable}{ll}
    \caption{Sample generation results of DORA on the interactive setting.}
    \label{table9} \\
        \hlineB{10}
        User: & i need information about a guesthouse called warkworth house. \\
        \hline Belief: & \{hotel-name: warkworth house, hotel-type: guesthouse\} \\
        \hline \multirow{2}{*}{Action:}
        & [booking]-[inform], [hotel]-[inform]-area, [hotel]-[inform]-name, [hotel]-[inform]-address, \\
        & [hotel]-[inform]-phone, [hotel]-[inform]-postcode \\
        \hline \multirow{2}{*}{Response:} 
        & [hotel\_name] is located in the [hotel\_area] at [hotel\_address] [hotel\_postcode]. their phone number \\
        & is [hotel\_phone]. would you like to book a room? \\
        \hlineB{5}
        User: & yes, book a room for two people. \\
        \hline Belief: & \{hotel-people: 2, hotel-name: warkworth house, hotel-type: guesthouse\} \\
        \hline Action: & [booking]-[request]-day, [booking]-[request]-stay \\
        \hline Response: & what day will you be arriving and how long will you be staying? \\
        \hlineB{5}
        User: & for two days from monday. \\
        \hline \multirow{2}{*}{Belief:} 
        & \{hotel-stay: 2, hotel-day: monday, hotel-people: 2, hotel-name: warkworth house, \\
        & hotel-type: guesthouse\} \\
        \hline Action: & [booking]-[book]-reference, [general]-[reqmore] \\
        \hline Response: 
        & booking was successful. reference number is : [hotel\_reference]. can i help with anything else? \\
        \hlineB{5}
        User: & i want to book a indian restaurant called curry prince. \\
        \hline \multirow{2}{*}{Belief:} 
        & \{restaurant-food: indian, restaurant-name: curry prince, hotel-stay: 2, hotel-day: monday, \\ 
        & hotel-people: 2, hotel-name: warkworth house, hotel-type: guesthouse\} \\
        \hline Action:
        & [booking]-[request]-people, [restaurant]-[inform]-name, [restaurant]-[inform]-address \\
        \hline Response: 
        & [restaurant\_name] is located at [restaurant\_address]. how many people will be dining? \\
        \hlineB{5}
        User: & book the restaurant for two people at 19:300 on monday. \\
        \hline \multirow{3}{*}{Belief:} 
        & \{restaurant-time: 19:30, restaurant-day: monday, restaurant-people: 2, restaurant-food: indian, \\
        & restaurant-name: curry prince, hotel-stay: 2, hotel-day: monday, hotel-people: 2, \\ 
        & hotel-name: warkworth house, hotel-type: guesthouse\} \\
        \hline Action: & [booking]-[book]-reference, [general]-[reqmore] \\
        \hline \multirow{2}{*}{Response:} 
        & booking was successful. the table will be reserved for 15 minutes. reference number \\
        & is : [restaurant\_reference]. is there anything else i can help you with? \\
        \hlineB{5}
        User: & can you give me the phone number? \\
        \hline \multirow{3}{*}{Belief:} 
        & \{restaurant-time: 19:30, restaurant-day: monday, restaurant-people: 2, restaurant-food: indian, \\
        & restaurant-name: curry prince, hotel-stay: 2, hotel-day: monday, hotel-people: 2, \\ 
        & hotel-name: warkworth house, hotel-type: guesthouse\} \\
        \hline Action: & [restaurant]-[inform]-phone, [restaurant]-[inform]-name \\
        \hline Response: & the phone number for [restaurant\_name] is [restaurant\_phone] \\
        \hlineB{5}
        User: & i also need a taxi from the guesthouse to the restaurant. \\
        \hline \multirow{4}{*}{Belief:} 
        & \{taxi-destination: curry prince, taxi-departure: warkworth house, restaurant-time: 19:30, \\ 
        & restaurant-day: monday, restaurant-people: 2, restaurant-food: indian, \\
        & restaurant-name: curry prince, hotel-stay: 2, hotel-day: monday, hotel-people: 2, \\ 
        & hotel-name: warkworth house, hotel-type: guesthouse\} \\
        \hline Action: & [taxi]-[request]-leave, [taxi]-[request]-arrive \\
        \hline Response: & what time would you like to leave or arrive by? \\
        \hlineB{5}
        User: & i will leave at 19:00. \\
        \hline \multirow{4}{*}{Belief:} 
        & \{taxi-leave: 19:00, taxi-destination: curry prince, taxi-departure: warkworth house, \\ 
        & restaurant-time: 19:30, restaurant-day: monday, restaurant-people: 2, restaurant-food: indian, \\
        & restaurant-name: curry prince, hotel-stay: 2, hotel-day: monday, hotel-people: 2, \\ 
        & hotel-name: warkworth house, hotel-type: guesthouse\} \\
        \hline Action: & [general]-[reqmore], [taxi]-[inform]-phone, [taxi]-[inform]-car \\
        \hline \multirow{2}{*}{Response:}
        & i have booked a [taxi\_car] for you. the contact number is [taxi\_phone]. is there anything else \\
        & i can help you with? \\
        \hlineB{5}
        User: & thank you for your help. have a nice day! \\
        \hline \multirow{4}{*}{Belief:} 
        & \{taxi-leave: 19:00, taxi-destination: curry prince, taxi-departure: warkworth house, \\ 
        & restaurant-time: 19:30, restaurant-day: monday, restaurant-people: 2, restaurant-food: indian, \\
        & restaurant-name: curry prince, hotel-stay: 2, hotel-day: monday, hotel-people: 2, \\ 
        & hotel-name: warkworth house, hotel-type: guesthouse\} \\
        \hline Action: & [general]-[bye], [general]-[welcome] \\
        \hline Response: & you're welcome! have a great day! \\
        \hlineB{10}
\end{longtable}

\begin{longtable}{ll}
    \caption{Another sample generation results of DORA on the interactive setting.}
    \label{table10} \\
        \hlineB{10}
        User: & i am looking for a cheap hotel in centre city. can you recommend one for me? \\
        \hline Belief: & \{hotel-area: centre, hotel-pricerange: cheap\} \\
        \hline \multirow{2}{*}{Action:}
        & [hotel]-[inform]-name, [hotel]-[inform]-address, [hotel]-[inform]-phone, \\
        & [hotel]-[inform]-postcode \\
        \hline \multirow{2}{*}{Response:}
        & i have [hotel\_name] located at [hotel\_address] [hotel\_postcode]. their phone number is \\
        & [hotel\_phone]. \\
        \hlineB{5}
        User: & i will stay alone. can you book the hotel from friday to sunday? \\
        \hline Belief: 
        & \{hotel-stay: 3, hotel-day: friday, hotel-people: 1, hotel-area: centre, hotel-pricerange: cheap\} \\
        \hline Action: & [booking]-[book]-reference, [booking]-[book]-name, [general]-[reqmore] \\
        \hline \multirow{2}{*}{Response:}
        & i have you booked at the [hotel\_name]. your reference number is [hotel\_reference]. can i \\
        & help with anything else? \\
        \hlineB{5}
        User: & i also need a italian restaurant in the same area. \\
        \hline \multirow{2}{*}{Belief:} 
        & \{restaurant-food: italian, restaurant-area: centre, hotel-stay: 3, hotel-day: friday, hotel-people: 1, \\ 
        & hotel-area: centre, hotel-pricerange: cheap\} \\
        \hline Action:
        & [restaurant]-[inform]-name, [restaurant]-[inform]-address, [restaurant]-[inform]-phone \\
        \hline Response:
        & [restaurant\_name] is located at [restaurant\_address] and their phone number is [restaurant\_phone]. \\
        \hlineB{5}
        User: & book the restaurant on 18:00 saturday for one \\
        \hline \multirow{3}{*}{Belief:} 
        & \{restaurant-time: 18:00, restaurant-day: saturday, restaurant-people: 1, restaurant-food: italian, \\
        & restaurant-area: centre, hotel-stay: 3, hotel-day: friday, hotel-people: 1, hotel-area: centre, \\ 
        & hotel-pricerange: cheap\} \\
        \hline Action: & [booking]-[book]-reference, [general]-[reqmore] \\
        \hline \multirow{2}{*}{Response:}
        & booking was successful. the table will be reserved for 15 minutes. reference number is : \\
        & [restaurant\_reference]. anything else i can help you with? \\
        \hlineB{5}
        User: & can you recommend a architecture for me in the same area? \\
        \hline \multirow{3}{*}{Belief:} 
        & \{restaurant-time: 18:00, restaurant-day: saturday, restaurant-people: 1, restaurant-food: italian, \\
        & restaurant-area: centre, hotel-stay: 3, hotel-day: friday, hotel-people: 1, hotel-area: centre, \\ 
        & hotel-pricerange: cheap, attraction-type: architecture, attraction-area: centre\} \\
        \hline \multirow{2}{*}{Action:} 
        & [attraction]-[recommend]-name, [attraction]-[recommend]-address, \\
        & [attraction]-[recommend]-postcode \\
        \hline \multirow{2}{*}{Response:}
        & i recommend [attraction\_name], it is located at [attraction\_address], postcode \\ 
        & [attraction\_postcode]. \\
        \hlineB{5}
        User: & that's all i need. thank you \\
        \hline \multirow{3}{*}{Belief:} 
        & \{restaurant-time: 18:00, restaurant-day: saturday, restaurant-people: 1, restaurant-food: italian, \\
        & restaurant-area: centre, hotel-stay: 3, hotel-day: friday, hotel-people: 1, hotel-area: centre, \\ 
        & hotel-pricerange: cheap, attraction-type: architecture, attraction-area: centre\} \\
        \hline Action: & [general]-[bye], [general]-[welcome] \\
        \hline Response: & you're welcome. have a great day! \\
        \hlineB{10}
\end{longtable}

\bibliographystyle{elsarticle-harv}

\end{document}